\documentclass{article}
\usepackage[utf8]{inputenc}
\usepackage{authblk}
\usepackage{adjustbox}

\usepackage{booktabs} % For formal tables
\usepackage{multirow} % For merged cell in tables
\usepackage{multicol}
\usepackage{diagbox} % For backslash column in tables
\usepackage{array}
\usepackage{comment}
\newcolumntype{x}[1]{>{\centering\arraybackslash}p{#1}}
\newcolumntype{y}[1]{>{\arraybackslash}p{#1}}

\usepackage{listings}

\usepackage{graphicx}
\usepackage{caption}
\usepackage{subcaption}

\usepackage{color}
\definecolor{green}{rgb}{0, 0.42, 0.24}
\definecolor{blue}{rgb}{0, 0.53, 0.74}

\usepackage{algorithm}% http://ctan.org/pkg/algorithms
\usepackage{algpseudocode}% http://ctan.org/pkg/algorithmicx

\usepackage{amsmath}
\DeclareMathOperator*{\argmax}{arg\,max}

\title{Man versus Machine: AutoML and Human Experts' Role in Phishing Detection}
\author[1]{Rizka Purwanto}
\author[1,2,3]{Arindam Pal}
\author[1]{Alan Blair}
\author[1]{Sanjay Jha}
\affil[1]{University of New South Wales, Australia}
\affil[2]{Data61, CSIRO, Sydney, Australia}
\affil[3]{Cyber Security Cooperative Research Centre, Australia}

\date{
    This work has been submitted to the IEEE for possible publication. Copyright may be transferred without notice, after which this version may no longer be accessible.
}

\begin{document}

\maketitle

\begin{abstract}
Machine learning (ML) has developed rapidly in the past few years and has successfully been utilized for a broad range of tasks, including phishing detection. However, building an effective ML-based detection system is not a trivial task, and requires data scientists with knowledge of the relevant domain. Automated Machine Learning (AutoML) frameworks have received a lot of attention in recent years, enabling non-ML experts in building a machine learning model. This brings to an intriguing question of whether AutoML can outperform the results achieved by human data scientists. Our paper compares the performances of six well-known, state-of-the-art AutoML frameworks on ten different phishing datasets to see whether AutoML-based models can outperform manually crafted machine learning models. Our results indicate that AutoML-based models are able to outperform manually developed machine learning models in complex classification tasks, specifically in datasets where the features are not quite discriminative, and datasets with overlapping classes or relatively high degrees of non-linearity. Challenges also remain in building a real-world phishing detection system using AutoML frameworks due to the current support only on supervised classification problems, leading to the need for labeled data, and the inability to update the AutoML-based models incrementally. This indicates that experts with knowledge in the domain of phishing and cybersecurity are still essential in the loop of the phishing detection pipeline.
\end{abstract}

\section{Introduction}
\label{sec:introduction}
Despite the availability of anti-phishing technologies, phishing attacks are still thriving and have caused data breaches of personal sensitive information and private company data. Phishing attacks were reported to double in 2020 \cite{apwg2021}, and have caused significant financial losses of roughly between \$60 million and \$3 billion per year in the United States \cite{Hong2012}. With the rapidly evolving nature of phishing, it would be ideal to have an automated detection system which could quickly adapt to phishing data changes and robustly detect these attacks.

Phishing is a cyber-attack that aims at stealing sensitive information by impersonating a legitimate person, company, or organization. By using social engineering and psychological tactics, phishing attackers exploit human vulnerabilities by sending messages that seem authentic and usually have a sense of urgency, persuading users to follow the attacker's instruction in the message. While phishing attacks nowadays are also advertised through voice messages and SMS messages, reports show that phishing is still dominated by the traditional ones delivered via emails \cite{apwg2021}.

\begin{comment}
\begin{itemize}
    \item besides its dynamic nature, the number of phishing websites are growing rapidly, which brings the urgency to detect it quickly without interrupting user's activity  on the web browsers
    \item it has been shown that machine learning could effectively detect phishing attacks
\end{itemize}
\end{comment}

%With AutoML frameworks, people without deep statistical or ML knowledge are able to build machine learning models to perform classification for a specific task. This provides a way for domain experts to focus more on the performing research on the problem which are to be tackled and the deployment of the machine learning models.
%\item while machine learning is beneficial, developing such models require human experts

During the past decade, machine learning has developed rapidly and has successfully performed a broad range of tasks. Machine learning techniques have shown to be effective in detecting phishing attacks. However, building a machine learning model is not a trivial task, and requires data scientists with knowledge of the relevant domain. In the past couple of years, Automated Machine Learning (AutoML) has gained much attention, enabling non-ML experts in building a machine learning model.

The emergence of AutoML frameworks brings us to the question of whether AutoML-generated models could outperform manually trained machine learning models on phishing data, how AutoML could assist non-experts in building machine learning models, and to what extent we could automate the whole process of building an ML-based detection system. There are a number of past studies that have tried to investigate whether the existence of AutoML frameworks would affect the roles of human experts in the ML development pipeline \cite{crisan_fits_2021}. However, to the best of our knowledge, we are the first to discuss this topic specifically in the case of phishing detection systems.

% This study aims to answer the following questions:
% \begin{itemize}
%     \item Can AutoML based classification models outperform manually crafted machine learning models on phishing datasets?
%     \item To what degree can we fully automate the  phishing detection process using AutoML?
%     \item What are the challenges of implementing phishing detection systems using AutoML?
%     % \item How do the features automatically selected by AutoML perform compared to those manually selected by data scientists and domain experts?
% \end{itemize}

\section{Automated Machine Learning}
\label{sec:automl}
% In this section, we will provide some literature review on past research on automated machine learning and the latest studies on techniques that automates the process of building machine learning models.

There has been significant research conducted in the areas of machine learning and deep learning since 1995, resulting in the development of various tools, such as Weka (1990s), scikit-learn (2007-2010), Tensorflow (2015), Keras (2015), and so on \cite{truong_towards_2019}. The emergence of these tools has enabled multidisciplinary research and the application of machine learning and deep learning techniques, which has shown promising results, demonstrating the potential of machine learning models to solve various problems. However, it has also become evident that developing a machine learning model is a complex task, requiring intuition, experience, and technical expertise to tune the model's hyperparameters. The heavy reliance of machine learning development on human experts motivates researchers to explore the possibility of inventing a technique to automate the development of machine learning models. Such attempts focusing on AutoML projects have been initiated by researchers and machine learning practitioners, and were followed by startups that sell AutoML frameworks as part of their business models. Some of the first AutoML tools are AutoWEKA \cite{kotthoff_auto-weka_2017} which is based on Weka \cite{Hall2009}, followed by auto-sklearn \cite{Feurer2015} and TPOT \cite{olson2016} which are both built on the scikit-learn library on Python \cite{scikit-learn}. Various AutoML frameworks have also emerged as the product of the ChaLearn AutoML challenge competitions \cite{automlChaLearn} between 2015 and 2018, which is still conducted every year until now.

%In general, Automated machine learning (AutoML) tools aim to automate the process of applying machine learning methods, which covers the complete machine learning (ML) pipeline from processing the raw dataset to building a deployable machine learning model. A common full ML pipeline is shown in Figure~\ref{fig:ml_pipeline}.

\begin{figure*}
    \centering
    \includegraphics[width=\textwidth]{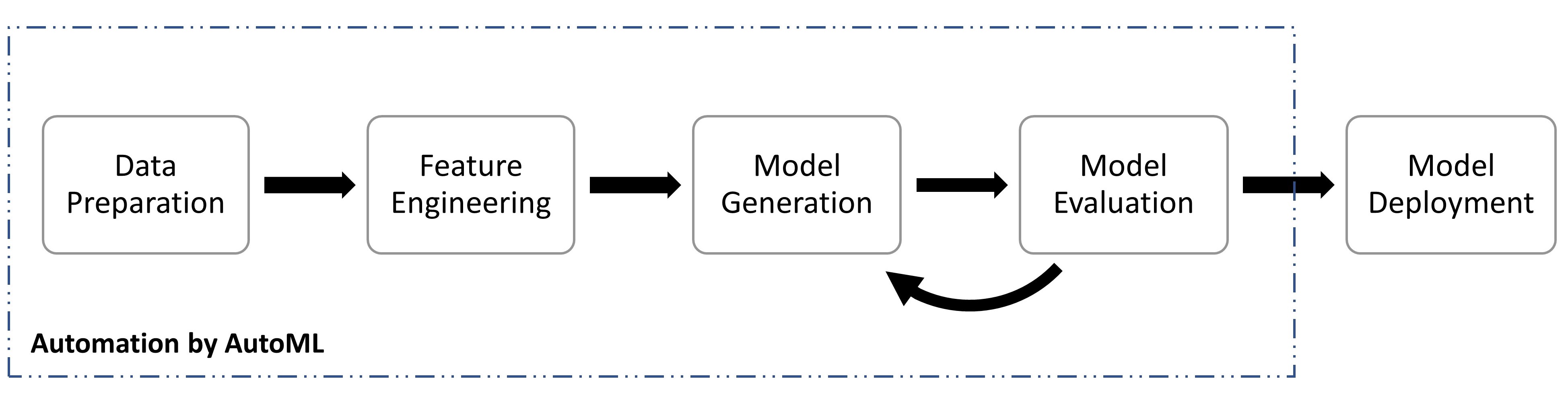}
    \caption{Standard Machine Learning Pipeline}
    \label{fig:ml_pipeline}
\end{figure*}

\color{black}
There are some existing studies which focus on performing a thorough literature review on the comparison and discussions on past works in AutoML approaches and tools \cite{he_automl_2021, waring_automated_2020}. In general, AutoML tools aim to automate various aspects of the machine learning pipeline, including data preprocessing, feature engineering, model training and validation. A standard full ML pipeline is shown in Figure~\ref{fig:ml_pipeline}. Based on previous literature reviews of AutoML tools \cite{waring_automated_2020, he_automl_2021}, we could divide AutoML frameworks based on which aspect of the machine learning pipeline it tries to automate; namely data preparation, feature engineering, model generation, and model evaluation.

\subsection{Data Preparation}
Data preparation includes data collection and preprocessing of the data, such as data cleaning and data augmentation. Automated data collection is necessary for tasks in which the data needs to be extended and continuously updated. Data cleaning aims to remove noises and handle missing values in the data. Meanwhile, data augmentation enhances the model's performance by generating new data based on the existing data and by preventing the trained model from over-fitting.

\subsubsection{Data Collection}
Plenty of open datasets are now being shared among researchers, especially image datasets (e.g. MNIST handwritten digital dataset \cite{lecun1998gradient}). However, it remains a challenge to obtain a high quality dataset for some specific tasks, including phishing websites and email datasets which require anonymization. This issue can be solved with two different approaches: data searching and data synthesis. Some automated methods that help in the performance of data searching include learning-based self-labeling methods for unlabeled data \cite{collins2008towards} and synthetic minority over-sampling technique (SMOTE) \cite{chawla2002smote} to deal with dataset class imbalances. Meanwhile, various data generation techniques also exist to automate data synthesis process, for example using Generative Adversarial Networks (GAN) \cite{karras2019style, oh2018learning, bowles2018gan} and reinforcement learning-based methods in data simulators \cite{ruiz2018learning}.

\subsubsection{Data Cleaning}
Data cleaning is an essential process to eliminate data noises, which could negatively affect machine learning model performances if not removed. However, data cleaning is generally a costly task, since it requires experts with specialist knowledge. Over time, there have been efforts to automate the process of data cleaning. Initially, data cleaning was performed through crowd-sourcing \cite{he_automl_2021}. To enhance efficiency, a past work proposed a data cleaning technique in which data scientists define specific data cleaning operations based on a small subset of the data, and apply these cleaning operations to the full dataset. Past studies attempted to further improve this method using machine learning techniques, such as boosting and hyperparameter optimization, to find the best data cleaning operation pipeline or combination \cite{krishnan2017boostclean, krishnan2019alphaclean}.  To be applicable for real world data, the data cleaning methods should be able to clean data steadily. Several past works have proposed a technique to evaluate data cleaning algorithms that can perform continuously, and to orchestrate cleaning workflows that can learn from past cleaning tasks.

%The data cleaning process is generally a costly task, since it requires experts with specialist knowledge. Over time, data cleaning efforts have been shifting from crowd-sourcing to automation \cite{he_automl_2021}. Some past studies have proposed clever methods to automate the process of data cleaning by using meta-learning techniques \cite{gemp2017automated}, and treating the task as a boosting problem \cite{krishnan2017boostclean} or hyperparameter optimization problem \cite{krishnan2019alphaclean}.

\subsubsection{Data Augmentation}
Data augmentation aims to enrich the dataset by generating new data by transforming the existing data. Past studies have proposed various methods to perform neural-based transformations on image data, such as adversarial noise \cite{mikolajczyk2018data}, neural style transfers \cite{mikolajczyk2019style}, and GAN-based techniques \cite{antoniou2017data}. Meanwhile, there are two approaches to textual data augmentation: data warping and synthetic over-sampling \cite{wong2016understanding}. Recently, various methods have been proposed to search for augmentation policies for different tasks using reinforcement learning \cite{cubuk2019autoaugment}, and various other improved algorithms \cite{lin2019online, lingchen2020uniformaugment}.

\begin{figure*}
    \centering
    \includegraphics[width=\textwidth]{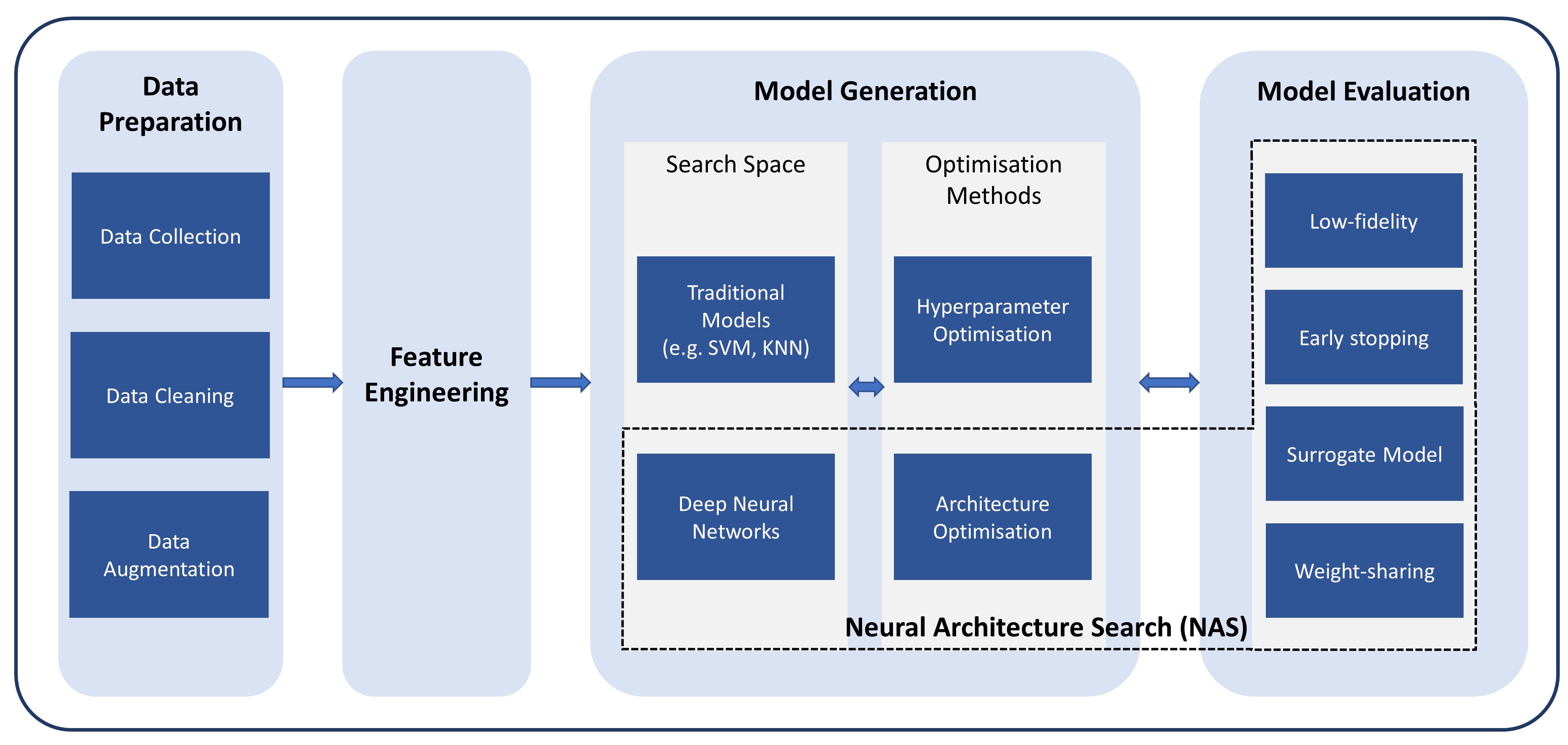}
    \caption{AutoML Pipeline and Components \cite{he_automl_2021}}
    \label{fig:automl_pipeline}
\end{figure*}

\subsection{Feature Engineering}
Feature engineering in supervised machine learning problems is defined as the process of finding explanatory variables that are predictive of the classification outcome \cite{waring_automated_2020}. This process is typically performed in a trial-and-error fashion and often requires extensive knowledge of the relevant domain. Feature engineering holds an important role, since its quality affects the machine learning model's performance heavily \cite{domingos2012few}; however, it is also a time-consuming task. To help with this process, some AutoML tools provided automated feature engineering methods with the goal of constructing new feature sets that give the best machine learning model performance.

The automated feature engineering task can be formally defined as follows \cite{waring_automated_2020}. Given a feature set with $m$ number of features, $F=[f_1, f_2, ..., f_m]$, a target vector, $y$, and a machine learning algorithm, $M$, let $P_M(F,y)$ indicate the performance of the model with the corresponding algorithm, target vector, and feature set. Assume there are $k$ transformation functions, $t_1, t_2, ..., t_k$ which can be applied to the features, and a sequence of transformations of the features $s=t_1(t_2...(f_1))$. The goal of the automated feature engineering process is to find a set of transformation sequences $S=[s_1, ..., s_r]$ to produce a new set of features, $F_{new}$, which satisfies $\argmax_{S}{P_M(F_{new},y)}$. The goal of automated feature engineering is to obtain the best set of features and feature transformations which give the best classification performance from structured data.

%The automated feature engineering methods differ from representation learning in deep learning methods.
%Meanwhile, representation learning aims to obtain useful feature spaces in unstructured data types, e.g. images, texts, and audios/videos.

Traditionally, new features are constructed manually by performing some standard transformations, such as standardization, normalization, or feature discretization. To improve the efficiency of such processes, automatic feature construction methods using decision trees \cite{gama2004functional, zheng1998comparison}, genetic algorithms \cite{vafaie1998evolutionary}, and annotation-based approaches \cite{sondhi2009feature}, have been proposed to search and evaluate the best combination of transformations.

Besides the construction of new features, feature engineering can also be performed by reducing the feature dimensionality to extract the most informative features and reduce redundancies in the feature set. This process is performed by applying mapping functions, such as principal component analysis (PCA) or linear discriminant analysis (LDA). In recent studies, there have been other improved methods proposed to perform feature extraction, e.g., using autoencoder-based algorithms \cite{meng2017relational} and unsupervised feature extraction methods \cite{irsoy2017unsupervised}.

\subsection{Model Generation and Evaluation}
There are two important elements in model generation: search space and the optimization method. The search space defines the structure and design principles of the machine learning models. Given a certain model, we could apply hyperparameter optimization (HPO), which aims to find the optimal training-related hyperparameters (e.g. learning rate), and architecture optimization (AO), to obtain the best hyperparameters associated with the model's structure or design (e.g., number of neighbors for $k$-NN, or number of layers or neural architectures for deep neural networks).

% , which can be divided into two categories: traditional ML models (e.g., $k$-NN and SVM) and neural architectures

Traditional hyperparameter optimization strategies usually do not make any assumptions about the search space. One of the simplest hyperparameter optimization methods is grid search; a brute-force method to find the best set of hyperparameters, given a finite set of values for each hyperparameter specified by users. Another simple alternative to grid search is random search, which relies on sampling from a user-specified set of hyperparameter values under a certain budget constraint. There is also another kind of hyperparameter optimization method which performs "optimization from samples" \cite{conn2009introduction}, e.g. particle swarm optimization (PSO) \cite{escalante2009particle} and evolutionary algorithms \cite{back1996evolutionary}, which are both inspired by biological behaviours. Meanwhile, Bayesian optimization has emerged as the most advanced hyperparameter optimization method used in AutoML frameworks. Bayesian optimization builds a probabilistic model, which maps different hyperparameter configurations to their performance with some degree of uncertainty.

Besides hyperparameter optimization, finding the best model architecture is also an important and non-trivial task when building a machine learning classifier. With the emerging research in neural networks in the past decade, neural architecture search (NAS) has gained great interest in the AutoML community. There are three essential components of a neural architecture search (NAS): the search space of neural architectures, architecture optimization, and model evaluation methods. An intuitive method to perform model evaluation is to assess model performance after training and the neural network has converged. However, this takes an extensive time and resources due to the amount of computation needed. Past studies have focused on finding methods to accelerate the model training and model evaluation process. The first approach is by using low-fidelity model evaluation by reducing the model size. Another technique that has been proposed is weight sharing, which can make the model training time faster by utilizing knowledge regarding weights of prior network architectures. Past studies also proposed the use of surrogate-based methods which could estimate the black-box function of a neural network model, making it easier to obtain the best model configuration and performance. Another proposed approach is early stopping, which was initially used in classical ML to prevent over-fitting. More recent studies have improved early stopping to perform computation on smaller set of data, making it faster to compute. We will not cover this topic in detail in our paper. However, more thorough discussions on studies on hyperparameter optimization and architecture optimization, especially neural architecture search (NAS), are covered by Waring et al. \cite{waring_automated_2020} and He at al. \cite{he_automl_2021} in their literature reviews.

\section{Phishing Detection Systems}
\label{sec:phishing_detection_system}
\begin{figure}
    \centering
    \includegraphics[width=0.7\textwidth]{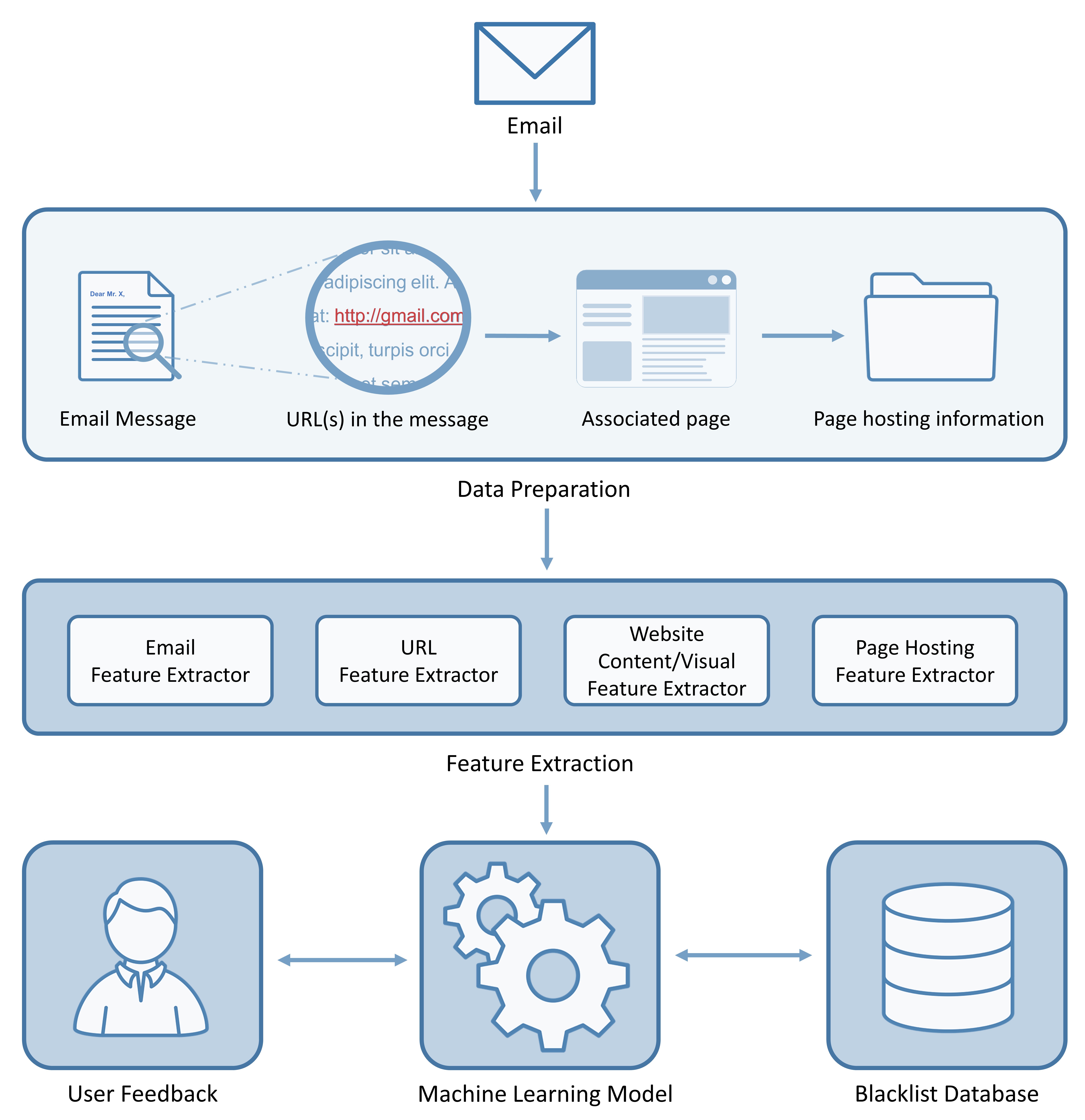}
    \caption{Phishing Detection Systems}
    \label{fig:phishing_detection_systems}
\end{figure}

Machine learning has shown to be effective in detecting phishing attacks based on past studies \cite{das2019sok}. While attackers have been using various strategies to conduct phishing attacks, emails still remain a primary delivery method for attackers. In this section, we provide the general workflow of phishing detection systems and how it interacts with external parties, e.g. users and blacklist providers. The phishing detection systems workflow is shown in Figure~\ref{fig:phishing_detection_systems}.

Phishing attacks start with the attacker broadcasting emails that contain a message which tries to convince the receivers to proceed further by clicking on the provided link. Automated phishing detection systems inspect this email's raw data and analyze the message, URL, content or visual appearance of the associated web page, and page hosting information. After automatically fetching this information, the feature extraction process is performed. Various past studies have focused on investigating the performance of features extracted from the phishing emails and websites. The output of the feature extraction process is a vector that contains values associated with each specific feature, each representing an email or website. Afterwards, a classifier is built, which predicts whether an email or a website is a phishing attack. This information is provided to users and phishing blacklist providers to update their database. In many cases, these detection systems also accept feedback from users or report from blacklist providers when misclassification occurs. This information is accepted as the ground truth for updating the machine learning model and improving its detection performance.

\section{Can AutoML outperform humans?}
\label{sec:can_automl_outperform_humans}

The advancement of AutoML frameworks raises the question of whether AutoML can outperform humans in building machine learning models for detecting phishing. To answer this question, we performed an experiment to compare the performances of the models built using AutoML and the ones that are manually crafted.

\subsection{Dataset and Performance Metrics}
To perform a comparison between AutoML and manually built machine learning models, we tested the models on various phishing email, URL, and website datasets. Besides using publicly available datasets \cite{fette2007learning, ma2009identifying, marchal2014phishstorm, mamun2016detecting, mohammad2012assessment}, we also performed feature extraction proposed in past studies \cite{fette2007learning, chandrasekaran2006phishing, gutierrez2018learning, ramanathan2012phishgillnet, verma2013semantic} on a raw phishing email dataset compiled by Verma et al. \cite{verma2019data} to construct new sets of phishing email data. In each \textit{task} of this experiment, we trained machine learning models using a specific dataset and compared the performance of models constructed with and without AutoML frameworks. Further details of each task are provided in Table~\ref{tab:dataset-details}.

\begin{table}[]
\centering
\begin{adjustbox}{width=1.3\textwidth,center}
%\resizebox{1.25\textwidth}{!}{%
\begin{tabular}{@{}lcccl@{}}
\toprule
\textbf{Task} & \textbf{Num of Rows} & \textbf{Num of Features} & \textbf{Num of Classes} & \multicolumn{1}{c}{\textbf{Details}} \\ \midrule
eml\_1a & 3668  & 3   & 2 & Raw phishing email dataset \cite{verma2019data} with features extraction proposed in \cite{fette2007learning}\\
eml\_1b & 3668  & 23  & 2 & Raw phishing email dataset \cite{verma2019data} with feature extraction proposed in \cite{chandrasekaran2006phishing}\\
eml\_1c & 3668  & 791 & 2 & Raw phishing email dataset \cite{verma2019data} with feature extraction proposed in \cite{gutierrez2018learning}\\
eml\_1d & 3668  & 10  & 2 & Raw phishing email dataset \cite{verma2019data} with feature extraction proposed in \cite{ramanathan2012phishgillnet}\\
eml\_1e & 3668  & 579 & 2 & Raw phishing email dataset \cite{verma2019data} with feature extraction proposed in \cite{verma2013semantic}\\
eml\_2    & 9205  & 3   & 2 & Phishing email dataset and feature extraction proposed in \cite{fette2007learning}\\
url\_1    & 96800 & 500 & 2 & Phishing website URL dataset and feature extraction proposed in \cite{ma2009identifying}\\
url\_2    & 76728 & 12  & 2 & Phishing website URL dataset and feature extraction proposed in \cite{marchal2014phishstorm}\\
url\_3    & 15185 & 79  & 5 & Phishing website URL dataset and feature extraction proposed in \cite{mamun2016detecting}\\
web\_1    & 8844  & 30  & 2 & Phishing website HTML dataset and feature extraction proposed in \cite{mohammad2012assessment} \\ \bottomrule
\end{tabular}%
%}
\end{adjustbox}
\caption{Phishing Dataset Details}
\label{tab:dataset-details}
\end{table}

While various metrics were measured during this experiment, we are particularly interested in the following performance metrics:
\begin{itemize}
    \item \textbf{Accuracy}\\
    Accuracy measures the number of correct predictions divided by the total size of the data, and can be expressed as:
    \begin{equation}
        \text{accuracy}(y,\hat{y})= \frac{1}{n_s} \sum_{i=0}^{n_s} 1\left(\hat{y}_{i}=y_{i}\right)
    \end{equation}
    
    where $\hat{y}_{i}$ is the predicted value of the $i$-th sample, $y_{i}$ is the corresponding true value, and $n_s$ is the total number of samples. In a multi-class classification setting, e.g. Task url\_{3}, this formula would calculate the subset accuracy, or the percentage of samples which are classified correctly.
    
    \item \textbf{AUC score}\\
    AUC, or Area Under the ROC Curve, is the total area under the ROC (receiver operating characteristic) curve, which would depict the model's performance in identifying the positive and negative examples for each class and finding the best threshold to separate between both examples. In a binary classification task, positive and negative examples would refer to phishing and legitimate samples. In a multi-class classification, we chose the \textit{one-vs-one} heuristic, where the dataset is split into one dataset for each class versus every other class \cite{hand2001simple}, and the final AUC score is obtaining by computing the average AUC of all possible pairwise combinations of classes.
    
    \item \textbf{F1-score}\\
    Due to the precision-recall trade-off, it is challenging to have both precision and recall high, especially in imbalanced datasets. F1-score calculates the harmonic mean of recall (true positive rate) and precision, measuring the model's performance correctly performing classification while heavily penalizing low recall or low precision scores.

    \item \textbf{Training duration}\\
    The training duration refers to the total amount of time each framework or algorithm takes to process the given dataset, until a classification model is produced. We do not include the time each model takes to perform prediction on the testing dataset, as it is generally negligible.
    
\end{itemize}

\subsection{Experiment Constraints and Setup}
In this study, we compared the performance of various mature open source AutoML frameworks, which are briefly described as follows:
\begin{itemize}
    \item \textbf{AutoGluon} \cite{agtabular}\\
    AutoGluon uses a multi-layer stack ensemble, in which multiple models are ensembled and stacked in multiple layers. The main difference between AutoGluon and existing AutoML frameworks is that AutoGluon utilizes almost every trained model to produce the final prediction instead of only selecting the best model.
    \item \textbf{auto-sklearn} \cite{NIPS2015_5872}\\
    Auto-sklearn was a winner of the ChaLearn AutoML Challenge 1 in 2015-2016 and ChaLearn AutoML Challenge 2 in 2017-2018. It uses Bayesian optimization to obtain the best machine learning pipeline. Auto-sklearn features automatic ensemble construction and uses meta-learning to increase the probability of finding a good pipeline by warm-starting the search procedure.
    \item \textbf{GAMA} \cite{Gijsbers2019}\\
    GAMA supports configurable AutoML pipelines, which allow the selection of optimization and post-processing algorithms. By default, GAMA searches over linear ML pipelines and creates a model ensemble in the post-processing step. These pipelines can be optimized with an asynchronous evolutionary algorithm or ASHA.
    \item \textbf{H2OAutoML} \cite{H2OAutoML20}\\
    H2OAutoML performs a random search, which is followed by a model stacking stage. This framework uses the H2O machine learning package by default, which supports distributed training.
    \item \textbf{hyperopt-sklearn} \cite{komer2014hyperopt}\\
    Hyperopt-sklearn allows various search strategies, including random search, and various sequential model based optimization (SMBO) techniques. These techniques include Tree of Parzen Estimators (TPE), Annealing and Gaussian Process Trees.
    \item \textbf{TPOT} \cite{OlsonGECCO2016}\\
    TPOT constructs machine learning pipelines of arbitrary length using scikit-learn algorithms \cite{scikit-learn} and allows the use of XGBoost algorithm. During its search, pre-processing and stacking are both considered. While the model's pipeline length is arbitrary, TPOT performs multi-objective optimization, in which it aims to keep the number of pipeline components minimal while optimizing the main selected metric. TPOT also provides support for sparse matrices, multiprocessing, and custom pipeline components.  
\end{itemize}

%  https://openml.github.io/automlbenchmark/automl_overview.html

To ensure that our evaluation was performed fairly, we trained and tested all the models using the OpenML AutoML Benchmarking Framework \cite{amlb2019} to make sure that the models were developed under the same constraints and with the same setup. Evaluating and comparing AutoML systems is challenging due to the subtle differences in problem definition, e.g. the design of the hyperparameter search space or the way time budgets are defined. The OpenML AutoML Benchmarking toolkit aims to address this issue by providing a standardized environment to perform in-depth experiments comparing a wide range of AutoML frameworks.

For each task in Table~\ref{tab:dataset-details}, we performed random splits on the whole dataset, then allocated 75\% for training and 25\% for testing respectively. All tasks were run using the same computing resources, with 8-core Intel Xeon and 32 GB RAM. The maximum runtime for each task was set to 3,600 seconds (1 hour). Each task could use a maximum of 8 cores when multiprocessing is available. To reduce biases resulting from outlier results, we repeated the experiment 10 times for each task and observed the consistency of the performance in the evaluation metrics. We selected the default metrics to optimize, which were the AUC score for binary classification tasks and log loss for multi-class classification tasks.

We also selected several traditional machine learning algorithms to compare with the models built using the aforementioned AutoML frameworks. The selected algorithms are Logistic Regression, SVM, KNN, Decision Tree, Random Forest, Multi-layer Perceptron, and Gaussian Naive Bayes, which are available in the \texttt{scikit-learn} Python package \cite{scikit-learn}. To train these models, we also used the same dataset split and computing resources as the one used in the AutoML setting, with a maximum of 8 core when multiprocessing was available. For each task in Table~\ref{tab:dataset-details}, we ran 10 experiments to obtain the best model for each algorithm and to observe any variance in the models' performance. We manually defined a set of model hyperparameters and performed random searches to obtain the best model, optimizing the AUC score in binary classification tasks and log loss in multi-class classification tasks. Unlike the AutoML experiment setting, we did not set the maximum runtime for each task as this feature was not available. However, we set the number of iterations during the random search for each task. The iteration number was manually set to 20 for SVM and 100 for all other algorithms.

\subsection{Results}
We computed the average performance metrics of each model and task in all experiments. To perform comparisons between models built using AutoML frameworks and non-AutoML algorithms, we selected two models from each task which would represent the best model built using the AutoML framework and non-AutoML algorithms. The best performance in this case would be in terms of the average AUC score as this metric was optimized during the model search process.

% Please add the following required packages to your document preamble:
% \usepackage{booktabs}
% \usepackage{graphicx}
% \begin{table}[btp!]
% \centering
% \begin{tabular}{@{}lcc@{}}
% \toprule
% \textbf{Task}    & \textbf{AUC Score} & \textbf{Accuracy} \\ \midrule
% \textbf{eml\_1a} & \textbf{0.135}              & \textbf{0.116}             \\
% \textbf{eml\_1b} & 0.005              & 0.002             \\
% \textbf{eml\_1c} & -0.008             & -0.001            \\
% \textbf{eml\_1d} & 0.002              & 0.007             \\
% \textbf{eml\_1e} & 0.027              & 0.003             \\
% \textbf{eml\_2}  & \textbf{0.233}              & \textbf{0.229}             \\
% \textbf{url\_1}  & -0.004             & -0.008            \\
% \textbf{url\_2}  & 0.002              & 0.006             \\
% \textbf{url\_3}  & 0.009              & 0.021             \\
% \textbf{web\_1}  & 0.01               & 0.013             \\ \bottomrule
% \end{tabular}
% \caption{Score Difference ($\text{best}_{AutoML}$-$\text{best}_{nonAutoML}$)}
% \label{tab:performance_diff}
% \end{table}

\begin{figure*}
    \centering
    \makebox[\textwidth][c]{\includegraphics[width=1.15\textwidth]{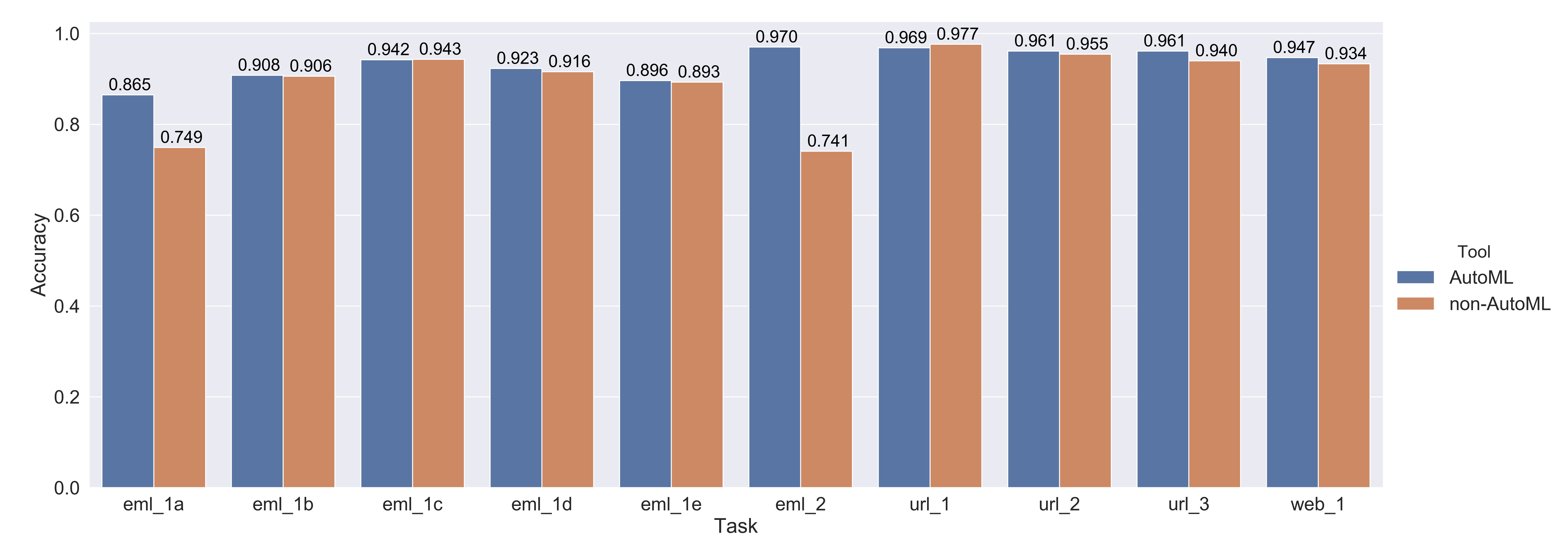}}%
    \caption{Accuracy}
    \label{fig:comparison_accuracy}
\end{figure*}

\begin{figure*}
    \centering
    \makebox[\textwidth][c]{\includegraphics[width=1.15\textwidth]{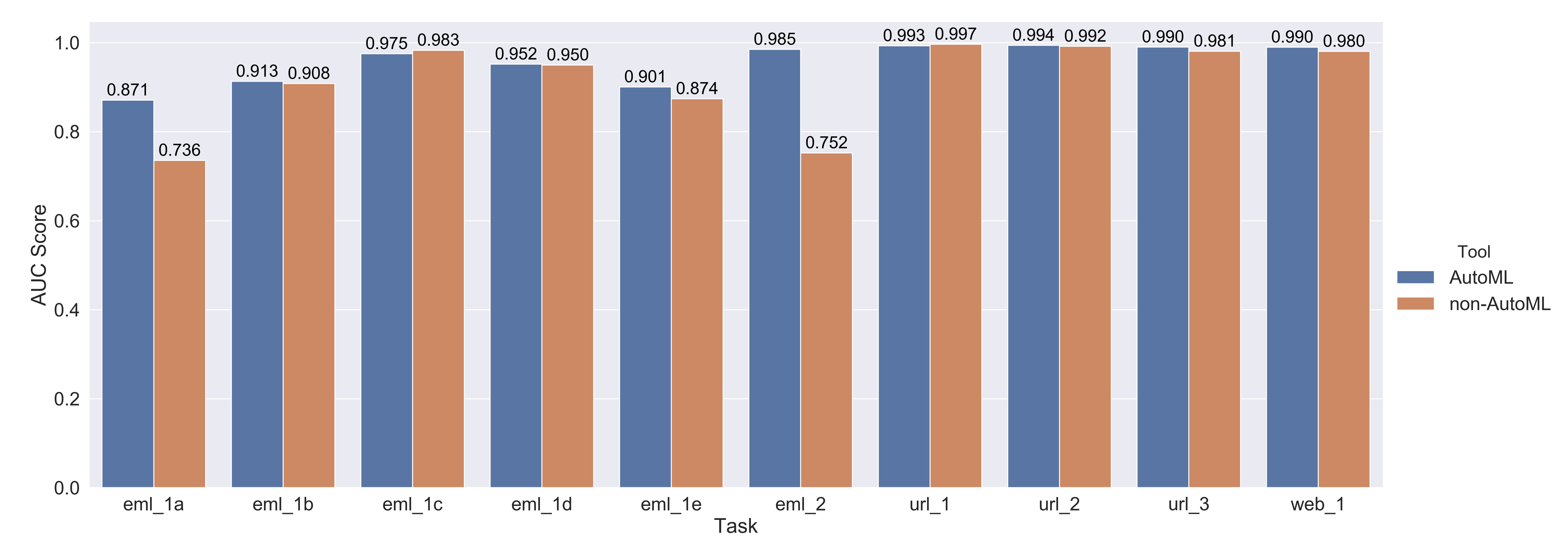}}%
    \caption{AUC Score}
    \label{fig:comparison_auc_score}
\end{figure*}

Comparisons of the accuracy and AUC score of the best model built using AutoML and non-AutoML frameworks are provided in Figure~\ref{fig:comparison_accuracy} and Figure~\ref{fig:comparison_auc_score}. As shown in these figures, the performances in terms of AUC score and accuracy are almost similar between models built using AutoML and manually developed ML models. However, there are some exceptions in Task eml\_1a and Task eml\_2 in which AutoML-based models significantly outperformed manually built models. The AUC score difference is between 13.5\% to 23.3\%, and the accuracy difference is between 11.6\% to 22.9\%.

\begin{figure*}
    \centering
    \makebox[\textwidth][c]{\includegraphics[width=1.15\textwidth]{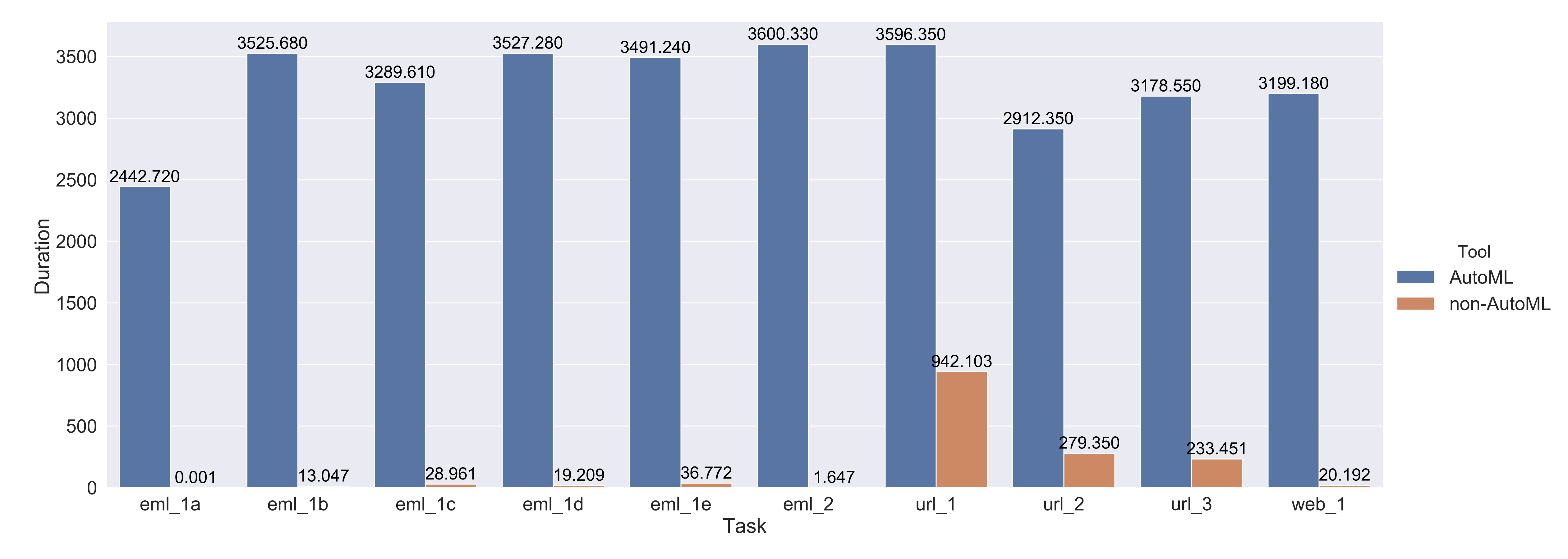}}%
    \caption{Duration}
    \label{fig:comparison_duration}
\end{figure*}

%To summarise these graphs better, we also provided the performance difference in Table~\ref{tab:performance_diff}. We consider a difference of 3\% and above as a significant score difference. Based on this assumption, it is shown that the performance in terms of AUC score and accuracy in all tasks are almost similar between models built using AutoML and manually crafted ML models. In some exceptions, AutoML models were able to significantly outperform non-AutoML models in Task eml\_1a and Task eml\_2 in terms of both the AUC score and accuracy. The score difference was roughly between 13.5\% to 23.3\%, and between 11.6\% to 22.9\% in terms of AUC Score and accuracy respectively.

In terms of training time, we found that it was much faster to train the models manually and achieve this level of performance without AutoML frameworks. Figure~\ref{fig:comparison_duration} provides a summary of the comparison of training duration between AutoML frameworks and traditional ML algorithms.

\section{When Does AutoML Outperform Humans?}
\label{sec:classification_task_complexity}

To gain a better understanding of the results in Section~\ref{sec:can_automl_outperform_humans}, we analyzed the complexity of performing classification on the datasets assigned to each task using the \texttt{DCoL} library \cite{dcol2010}. The aim of this experiment is to understand in what kind of classification task AutoML frameworks provide better results.

The complexity measures that are computed can be grouped into several categories, which are briefly described as follows.
\begin{itemize}
    \item \textbf{Measures of the overlaps from different classes based on the discriminative power of the features}, including the maximum Fisher's discriminant ratio (F1), directional-vector maximum Fisher's discriminant Ratio (F1v), volume of per-class bounding boxes overlap (F2), maximum individual feature efficiency (F3), and collective feature efficiency (F4).
    \item \textbf{Measures of linearity}, which includes the minimised sum of the error distance of a linear classifier (L1), training error of a linear classifier (L2), and non-linearity of a linear classifier (L3).
    \item \textbf{Neighborhood measures}, which includes the fraction of points on the class boundary (N1), ratio of intra/inter class nearest neighbor distance (N2), leave-one-out error rate of the one-nearest neighbor classifier (N3), non-linearity of the one-nearest neighbor classifier (N4), and maximum covering spheres fraction (T1).
    \item \textbf{Measures of dimensionality}, which includes the average number of points per dimension (T2).
    \item \textbf{Measures of class imbalance}, which includes the entropy of class proportions (C1) and imbalance ratio (C2).
\end{itemize}

\begin{figure*}
    \begin{minipage}[b]{\textwidth}
    \centering
    \makebox[\textwidth][c]{\includegraphics[width=1.5\textwidth]{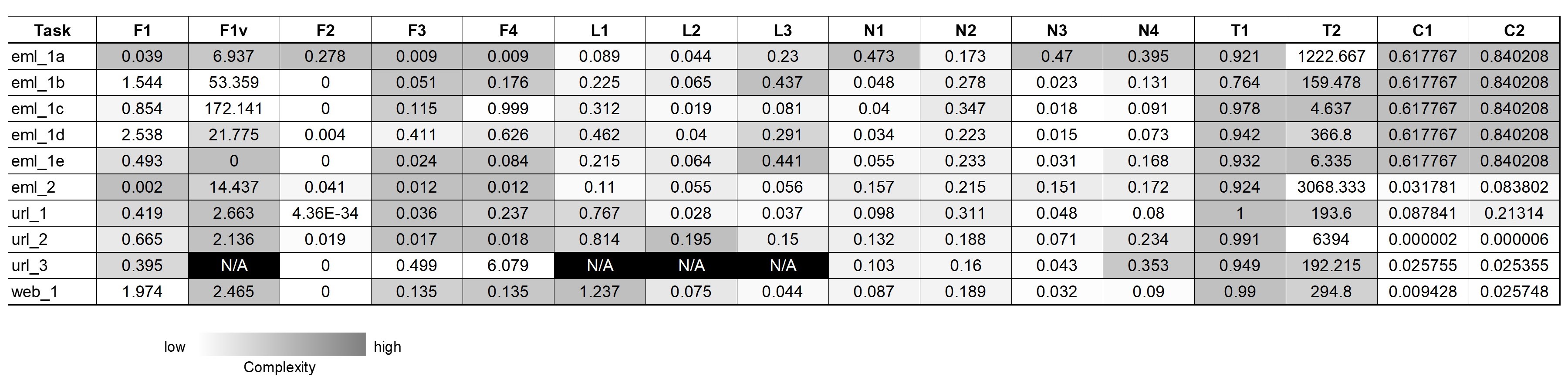}}%
    \end{minipage}

    \hfill

    \begin{minipage}[b]{\textwidth}
    \centering

    \resizebox{\textwidth}{!}{%
        \begin{tabular}{|l|lllll}
        \textbf{F1} &
          \multicolumn{1}{l|}{Maximum Fisher's discriminant ratio} &
          \multicolumn{1}{l|}{\textbf{L2}} &
          \multicolumn{1}{l|}{Training error of a linear classifier (linear SMO)} &
          \multicolumn{1}{l|}{\textbf{N4}} &
          Non-linearity of the one-nearest neighbor classifier \\
        \textbf{F1v} &
          \multicolumn{1}{l|}{Directional-vector maximum Fisher's discriminant ratio} &
          \multicolumn{1}{l|}{\textbf{L3}} &
          \multicolumn{1}{l|}{Non-linearity of a linear classifier (linear SMO)} &
          \multicolumn{1}{l|}{\textbf{T1}} &
          Fraction of maximum covering spheres \\
        \textbf{F2} &
          \multicolumn{1}{l|}{Overlap of the per-class bounding boxes} &
          \multicolumn{1}{l|}{\textbf{N1}} &
          \multicolumn{1}{l|}{Fraction of points on the class boundary} &
          \multicolumn{1}{l|}{\textbf{T2}} &
          Average number of points per dimension \\
        \textbf{F3} &
          \multicolumn{1}{l|}{Maximum (individual) feature efficiency} &
          \multicolumn{1}{l|}{\textbf{N2}} &
          \multicolumn{1}{l|}{Ratio of average intra/inter class nearest neighbor distance} &
          \multicolumn{1}{l|}{\textbf{C1}} &
          Entropy of Class Proportions \\
        \textbf{F4} &
          \multicolumn{1}{l|}{Collective feature efficiency (sum of each feature efficiency)} &
          \multicolumn{1}{l|}{\textbf{N3}} &
          \multicolumn{1}{l|}{Leave-one-out error rate of the one-nearest neighbor classifier} &
          \multicolumn{1}{l|}{\textbf{C2}} &
          Imbalance Ratio \\
        \textbf{L1} & Minimized sum of the error distance of a linear classifier (linear SMO) & \textbf{} &  & \textbf{} & 
        \end{tabular}%
        }

    \end{minipage}
    
    \caption{Classification Task Complexity}
    \label{tab:task_complexity}
\end{figure*}

A complete analysis of task complexity is provided in Figure~\ref{tab:task_complexity}. In this figure, darker cells are associated with higher complexity. In general, a higher parameter value usually indicates that the classification task is more complex, with an exception for F1, F1v, F3, F4, and T2 complexity measures where a higher measure value corresponds to simpler classification tasks. Higher complexity means that it is more difficult to achieve good classification performances. Further details on complexity measures are not discussed in our paper. However, we refer to \cite{ho2002complexity, ho2006measures, lorena2019complex} for those interested in more detailed explanations regarding measuring supervised classification complexity.

To observe the relationship between a task's complexity and classification performance, we performed a correlation test between AutoML performance gain or improvement ($\texttt{performance}_{AutoML}-\texttt{performance}_{nonAutoML}$) and a complexity measure based on one of the parameters previously mentioned. Full results of this analysis is provided in Table~\ref{tab:correlation_gain} (Appendix~\ref{app:correlation_analysis}). A summary of the correlation test with statistically significant results ($p<0.05$) is shown in Table~\ref{tab:correlation_gain_significant}.

Based on the correlation test, there are negative relationships between the F1 complexity measure (maximum Fisher's discriminant ratio) and the performance gain in terms of AUC score. Note that a higher Fisher's discriminant ratio would correspond to a simpler classification task. The negative relationship between F1 complexity measure and the AUC score gain would indicate that there would be a larger AUC score difference between AutoML-based models and manually built (non-AutoML) models in more complex tasks (with lower F1 complexity measure). The same goes with F3 and F4 complexity measures, in which both would have higher values when the classification task is simpler. Thus, a negative correlation between the F3 complexity measure and AUC score gain would indicate that AutoML-based models would outperform more significantly in more complex classification tasks (with lower F3 complexity measure). The correlation test also shows that AutoML-based models would outperform manually built models, in terms of accuracy, in more complex classification tasks with lower F4 complexity measure.

Furthermore, the correlation test also showed a positive correlation between some of the neighbor-based complexity measures (N1, N3, N4) and the AUC score performance gain. Unlike the previous complexity measures, a higher neighbor-based complexity measure indicates a more complex classification task. The results in Table~\ref{tab:correlation_gain_significant} show that the AutoML-based models would be more likely to outperform manually developed classification models in a more complex classification task with higher N1, N3, and/or N4 complexity measures. These results show that AutoML-based models would be most beneficial when used to build classification models in complex settings where the features are not quite discriminative, and in datasets with overlapping classes or relatively high degrees of non-linearity.

% Please add the following required packages to your document preamble:
% \usepackage{booktabs}
% \usepackage{graphicx}
\begin{table}[]
\centering
\resizebox{0.7\textwidth}{!}{%
\begin{tabular}{@{}cccc@{}}
\toprule
\textbf{Complexity Measure} & \textbf{Performance Gain} & \textbf{Correlation} & \textbf{p-value} \\ \midrule
F1 & AUC Score & -0.70909 & 0.021666 \\
F3 & AUC Score & -0.67273 & 0.033041 \\
F4 & Accuracy  & -0.70909 & 0.021666 \\
N1 & AUC Score & 0.745455 & 0.01333  \\
N3 & AUC Score & 0.721212 & 0.018573 \\
N4 & AUC Score & 0.769697 & 0.009222 \\ \bottomrule
\end{tabular}%
}
\caption{Correlation between Complexity Measure and Performance Gain ($p<0.05$)}
\label{tab:correlation_gain_significant}
\end{table}

Referring back to the results in Section~\ref{sec:can_automl_outperform_humans}, we found that this result is consistent with empirical findings. As shown in Figure~\ref{fig:comparison_accuracy} and Figure~\ref{fig:comparison_auc_score}, AutoML frameworks significantly outperformed manually crafted ML models in Task eml\_1a and Task eml\_2. As shown in Figure~\ref{tab:task_complexity}, Task eml\_1a is deemed to be complex as indicated by its feature-based complexity measures (F1, F3, F4) and neighborhood complexity measures (N1, N3, N4). Meanwhile, the complexity of Task eml\_2 is indicated by the feature-based measures (F1, F3, F4). This confirms that AutoML-based models outperform manually built ML models in these types of complex classification tasks.

\section{Automating Phishing Detection with AutoML Frameworks}
\label{sec:automl_for_phishing_detection}

In this section, we discuss further on the AutoML-based models' performances, followed by a discussion on the opportunity and challenges of the use of AutoML frameworks in automated phishing detection systems, and a highlight on the study limitations and potential future works.

\subsection{AutoML-based Models' Performances}
\begin{figure*}[]
    \centering
    \makebox[\textwidth][c]{\includegraphics[width=1.35\textwidth]{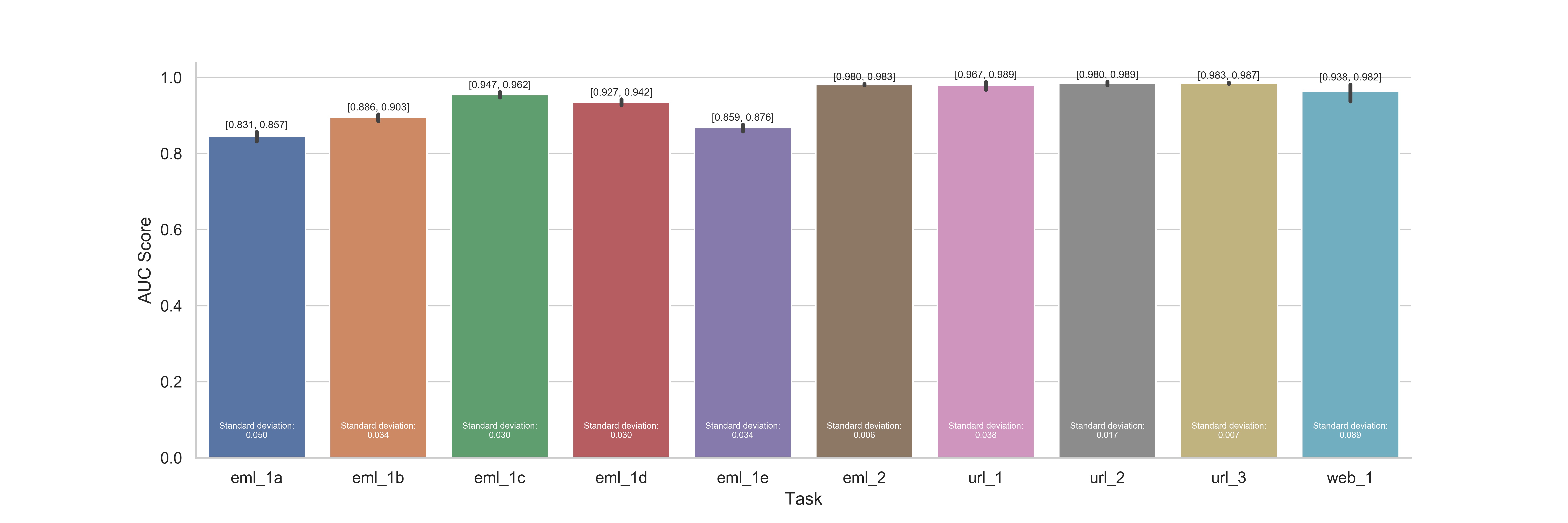}}%
    \caption{AutoML-based Models' AUC Score per Task}
    \label{fig:automl_all_auc}
\end{figure*}

\begin{figure*}[]
    \centering
    \makebox[\textwidth][c]{\includegraphics[width=1.35\textwidth]{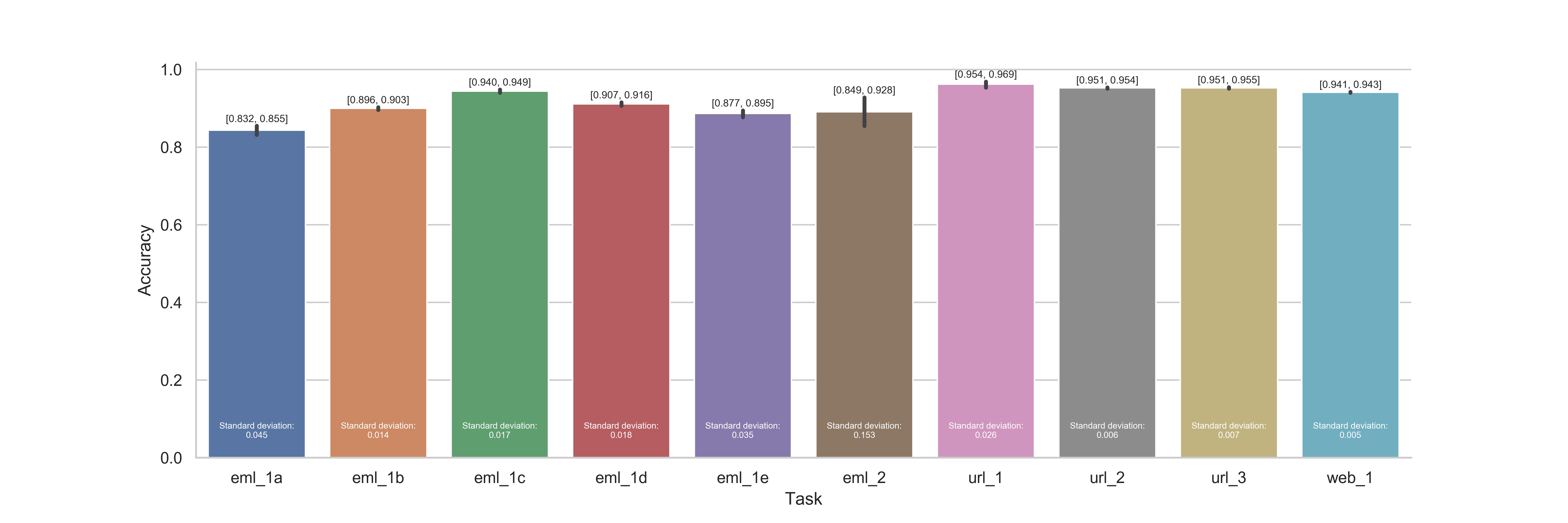}}%
    \caption{AutoML-based Models' Accuracy per Task}
    \label{fig:automl_all_acc}
\end{figure*}

\begin{figure*}[]
    \centering
    \makebox[\textwidth][c]{\includegraphics[width=1.35\textwidth]{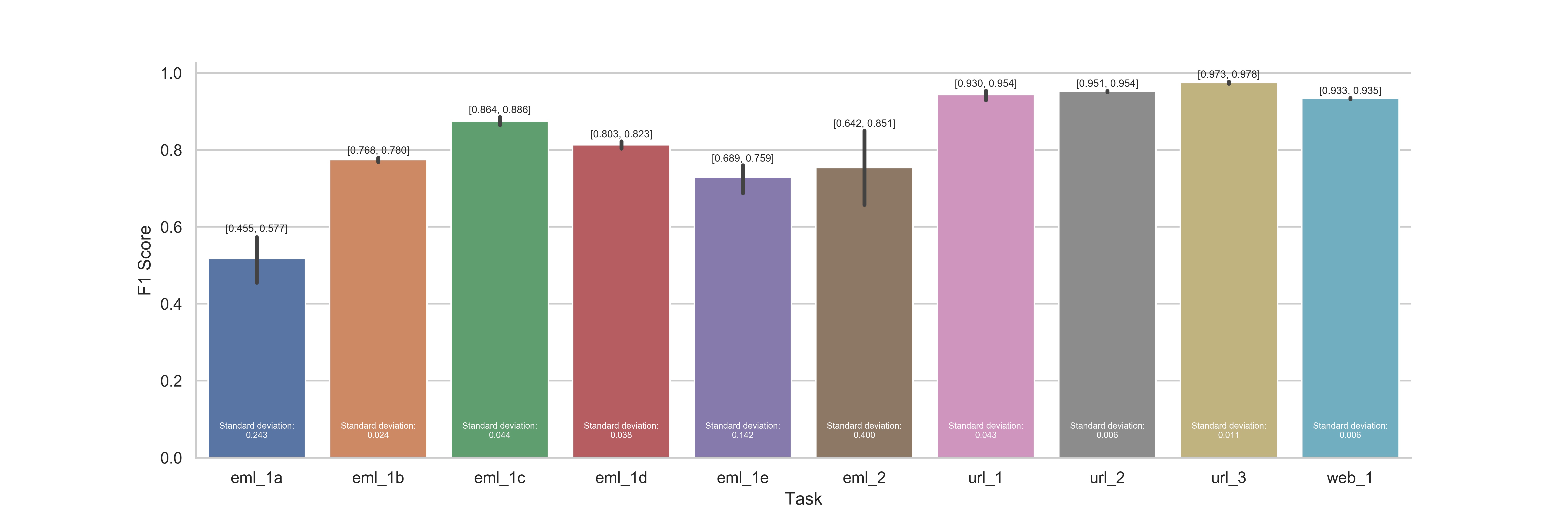}}%
    \caption{AutoML-based Models' F1 Score per Task}
    \label{fig:automl_all_f1_score}
\end{figure*}

In this section, we analyze further the AutoML-based models' performances in each classification task, and observe the relationship between a model's performance and complexity level of the classification task. We provide more details on the average accuracy, AUC score, and duration of each framework in Table~\ref{tab:automl_performance} in Appendix~\ref{app:model_perform}. The average AUC score, accuracy, and F1 score of each task is provided in Figure~\ref{fig:automl_all_auc}, Figure~\ref{fig:automl_all_acc}, and Figure~\ref{fig:automl_all_f1_score}. In these figures, we also computed the standard deviation and the confidence interval of each metric in every classification task.

We performed a correlation test to observe the relationship between the performance of an AutoML-based model on a specific classification task and a measure indicating the task's complexity, as shown in Table~\ref{tab:correlation_automl} (Appendix~\ref{app:correlation_analysis}). We provide a summary of the correlation analysis with statistically significant results in Table~\ref{tab:correlation_automl_significant}, As shown in this table, there are five complexity measures that have a significant correlation with AutoML-based models' performances, namely F4, L3, T1, C1, and C2 complexity measures. A higher F4 complexity measure would indicate a simpler classification task, whereas higher L3, T1, C1, or C2 complexity measure indicates that the classification task is more complex. The results in Table~\ref{tab:correlation_automl_significant} are quite intuitive in terms of the relationship between F4, L3, C1, C2 complexity measure and the performance metrics, where the performance would improve in simpler classification tasks. Interestingly, there is a positive correlation between AutoML-based model performance (accuracy and F1 score) and T1 complexity measure, indicating that the model seems to have better performance when classifying on datasets with higher fraction of maximum covering spheres.
%Referring back to the graphs in Figure~\ref{fig:automl_all_auc} and Figure~\ref{fig:automl_all_acc}, high AUC score often coincides with high accuracy, indicating that optimizing AUC score during model training with AutoML frameworks usually lead to a decent performance in terms of accuracy. An interesting case to observe further would be the results of Task eml\_2, which shows a relatively high AUC score of roughly 98\% with a significantly lower accuracy of around 89\%. This occurrence would be due to high complexity of Task eml\_2's in terms of the dataset's feature efficiency as demonstrated by its low F4 complexity measure, which leads to a lower accuracy based on the correlation test result in Table~\ref{tab:correlation_automl_significant}.

%Furthermore, Figure~\ref{fig:automl_all_auc} and Figure~\ref{fig:automl_all_f1_score} provide the results of Task url\_1, Task url\_2, Task url\_3, and Task web\_1 which consistently show a high AUC score and F1 score. On the other hand, there is more than 5\% difference between the AUC score and F1 score in Task eml\_1a, Task eml\_1b, Task eml\_1c, Task eml\_1d, Task eml\_1e, and Task eml\_2. Meanwhile, the results of Task eml\_1a, Task eml\_1b, Task eml\_1c, Task eml\_1d, and Task eml\_1e show significantly high C1 and C2 scores, which both have negative correlation with AutoML-based models' performances in terms of F1 score (Table~\ref{tab:correlation_automl_significant}).

% Please add the following required packages to your document preamble:
% \usepackage{booktabs}
% \usepackage{graphicx}
\begin{table}[]
\centering
\resizebox{0.7\textwidth}{!}{%
\begin{tabular}{@{}cccc@{}}
\toprule
\textbf{Complexity Measure} & \textbf{Performance Metric} & \textbf{Correlation} & \textbf{p-value} \\ \midrule
F4 & Accuracy  & 0.648485 & 0.04254  \\
L3 & Accuracy  & -0.69697 & 0.025097 \\
L3 & F1 Score  & -0.68485 & 0.028883 \\
L3 & AUC Score & -0.7697  & 0.009222 \\
T1 & Accuracy  & 0.830303 & 0.00294  \\
T1 & F1 Score  & 0.781818 & 0.007547 \\
C1 & F1 Score  & -0.69176 & 0.026678 \\
C1 & AUC Score & -0.834   & 0.002705 \\
C2 & F1 Score  & -0.73055 & 0.016409 \\
C2 & AUC Score & -0.88572 & 0.000649 \\ \bottomrule
\end{tabular}%
}
\caption{Correlation between Complexity Measure and AutoML-based Models' Performance ($p<0.05$)}
\label{tab:correlation_automl_significant}
\end{table}

\subsection{Opportunity and Challenges}
Machine learning models built using AutoML frameworks have shown to outperform manually built models (Figure~\ref{fig:comparison_accuracy} and Figure~\ref{fig:comparison_auc_score}). It is also shown that performance gain correlates with several complexity measures, such as F1, F3, F4, N1, N3, and N4 complexity measures (Table~\ref{tab:correlation_gain_significant}), indicating that human experts would benefit AutoML frameworks greatly in automating machine learning development, when dealing with complex classification tasks particularly with dataset features having low discriminative power, datasets with high class overlap, and dataset with high degree of non-linearity.

However, several challenges remain to be addressed in using AutoML frameworks for implementing a full machine learning pipeline of phishing detection systems. AutoML frameworks currently focus on supervised classification tasks with labeled datasets, which are not always available in the case of phishing classification tasks. Furthermore, most AutoML frameworks build stacked or ensembles of ML models, which are difficult to implement and update in an incremental learning setting. With these limitations, the role of data scientists and security experts is still crucial in observing whether the classification models need retraining due to the existence of concept drift, or if the features in the dataset are no longer representative of the attacks. It is also shown in Figure~\ref{fig:comparison_accuracy} and Figure~\ref{fig:comparison_auc_score} that while AutoML built models were able to outperform manually built ML models, the performance differences are not significant, except in several complex cases. Furthermore, the time needed to train manually built ML models is significantly lower when compared with training AutoML-based models (Figure~\ref{fig:comparison_duration}).

% \section{Limitations and Future Works}

\section{Conclusion}
\label{sec:conclusion}
Our study shows that AutoML frameworks are able to consistently produce models that perform similarly or better than manually built ML models. However, challenges remain in implementing these frameworks to fully automate phishing detection due to the support provided only on supervised classification problems, which consequently lead to the need for labeled data, and the inability to update the generated models incrementally. This indicates that experts with domain knowledge are still needed in-the-loop of the full phishing detection pipeline. Despite its limitations, our study also reveals that AutoML frameworks are able to outperform manually developed machine learning models in certain complex classification tasks, indicating that there are opportunities for utilizing AutoML to improve human-guided phishing detection systems. Future studies which further explore the collaboration of human experts and AutoML, as well as the design of the human-in-the-loop system architectures would be beneficial in improving phishing detection.

%%
%% The acknowledgments section is defined using the "acks" environment
%% (and NOT an unnumbered section). This ensures the proper
%% identification of the section in the article metadata, and the
%% consistent spelling of the heading.
\section*{Acknowledgement}
This work has been supported by the Cyber Security Cooperative Research Centre Limited, whose activities are partially funded by the Australian Government's Cooperative Research Centres Programme.

Rizka Widyarini Purwanto was supported by a UNSW University International Postgraduate Award (UIPA) scholarship. Any opinions, findings, and conclusions or recommendations expressed in this paper are those of the authors and do not necessarily reflect the views of the scholarship provider.

A sincere thank you to Muhammad Johan Alibasa for his constructive feedback and discussions on the manuscript.

\bibliographystyle{plain}
\bibliography{reference}

\begin{thebibliography}{10}

\bibitem{automlChaLearn}
{AutoML@ChaLearn}.
\newblock https://automl.chalearn.org/.
\newblock Accessed: 2021-03-29.

\bibitem{antoniou2017data}
Antreas Antoniou, Amos Storkey, and Harrison Edwards.
\newblock Data augmentation generative adversarial networks.
\newblock {\em arXiv preprint arXiv:1711.04340}, 2017.

\bibitem{apwg2021}
APWG.
\newblock {\em APWG Phishing Trends Reports}.
\newblock Anti Phishing Working Group, 2021.

\bibitem{back1996evolutionary}
Thomas Back.
\newblock {\em Evolutionary algorithms in theory and practice: evolution
  strategies, evolutionary programming, genetic algorithms}.
\newblock Oxford university press, 1996.

\bibitem{bowles2018gan}
Christopher Bowles, Liang Chen, Ricardo Guerrero, Paul Bentley, Roger Gunn,
  Alexander Hammers, David~Alexander Dickie, Maria~Vald{\'e}s Hern{\'a}ndez,
  Joanna Wardlaw, and Daniel Rueckert.
\newblock Gan augmentation: Augmenting training data using generative
  adversarial networks.
\newblock {\em arXiv preprint arXiv:1810.10863}, 2018.

\bibitem{chandrasekaran2006phishing}
Madhusudhanan Chandrasekaran, Krishnan Narayanan, and Shambhu Upadhyaya.
\newblock {Phishing Email Detection Based on Structural Properties}.
\newblock In {\em NYS Cyber Security Conference}, volume~3, 2006.

\bibitem{chawla2002smote}
Nitesh~V Chawla, Kevin~W Bowyer, Lawrence~O Hall, and W~Philip Kegelmeyer.
\newblock Smote: synthetic minority over-sampling technique.
\newblock {\em Journal of artificial intelligence research}, 16:321--357, 2002.

\bibitem{collins2008towards}
Brendan Collins, Jia Deng, Kai Li, and Li~Fei-Fei.
\newblock Towards scalable dataset construction: An active learning approach.
\newblock In {\em European conference on computer vision}, pages 86--98.
  Springer, 2008.

\bibitem{conn2009introduction}
Andrew~R Conn, Katya Scheinberg, and Luis~N Vicente.
\newblock {\em Introduction to derivative-free optimization}.
\newblock SIAM, 2009.

\bibitem{crisan_fits_2021}
Anamaria Crisan and Brittany Fiore-Gartland.
\newblock Fits and {Starts}: {Enterprise} {Use} of {AutoML} and the {Role} of
  {Humans} in the {Loop}.
\newblock {\em arXiv:2101.04296 [cs]}, January 2021.
\newblock arXiv: 2101.04296.

\bibitem{cubuk2019autoaugment}
Ekin~D Cubuk, Barret Zoph, Dandelion Mane, Vijay Vasudevan, and Quoc~V Le.
\newblock Autoaugment: Learning augmentation strategies from data.
\newblock In {\em Proceedings of the IEEE/CVF Conference on Computer Vision and
  Pattern Recognition}, pages 113--123, 2019.

\bibitem{das2019sok}
Avisha Das, Shahryar Baki, Ayman El~Aassal, Rakesh Verma, and Arthur Dunbar.
\newblock {SoK: A Comprehensive Reexamination of Phishing Research From the
  Security Perspective}.
\newblock {\em IEEE Communications Surveys \& Tutorials}, 22(1), 2019.

\bibitem{domingos2012few}
Pedro Domingos.
\newblock {A Few Useful Things to Know About Machine Learning}.
\newblock {\em Communications of the ACM}, 55(10):78--87, 2012.

\bibitem{agtabular}
Nick Erickson, Jonas Mueller, Alexander Shirkov, Hang Zhang, Pedro Larroy,
  Mu~Li, and Alexander Smola.
\newblock {AutoGluon-Tabular: Robust and Accurate AutoML for Structured Data}.
\newblock {\em arXiv preprint arXiv:2003.06505}, 2020.

\bibitem{escalante2009particle}
Hugo~Jair Escalante, Manuel Montes, and Luis~Enrique Sucar.
\newblock Particle swarm model selection.
\newblock {\em Journal of Machine Learning Research}, 10(2), 2009.

\bibitem{fette2007learning}
Ian Fette, Norman Sadeh, and Anthony Tomasic.
\newblock {Learning to Detect Phishing Emails}.
\newblock In {\em Proceedings of the 16th International Conference on World
  Wide Web}, 2007.

\bibitem{NIPS2015_5872}
Matthias Feurer, Aaron Klein, Katharina Eggensperger, Jost Springenberg, Manuel
  Blum, and Frank Hutter.
\newblock {Efficient and Robust Automated Machine Learning}.
\newblock In {\em Advances in Neural Information Processing Systems 28}, pages
  2962--2970. Curran Associates, Inc., 2015.

\bibitem{Feurer2015}
Matthias Feurer, Aaron Klein, Katharina Eggensperger, Jost~Tobias Springenberg,
  Manuel Blum, and Frank Hutter.
\newblock Efficient and robust automated machine learning.
\newblock NIPS'15, page 2755–2763, Cambridge, MA, USA, 2015. MIT Press.

\bibitem{gama2004functional}
Jo{\~a}o Gama.
\newblock Functional trees.
\newblock {\em Machine Learning}, 55(3):219--250, 2004.

\bibitem{amlb2019}
P.~Gijsbers, E.~LeDell, S.~Poirier, J.~Thomas, B.~Bischl, and J.~Vanschoren.
\newblock {An Open Source AutoML Benchmark}.
\newblock {\em arXiv preprint arXiv:1907.00909 [cs.LG]}, 2019.
\newblock Accepted at AutoML Workshop at ICML 2019.

\bibitem{Gijsbers2019}
Pieter Gijsbers and Joaquin Vanschoren.
\newblock {GAMA: Genetic Automated Machine learning Assistant}.
\newblock {\em Journal of Open Source Software}, 4(33):1132, 2019.

\bibitem{gutierrez2018learning}
Christopher~N Gutierrez, Taegyu Kim, Raffaele Della~Corte, Jeffrey Avery, Dan
  Goldwasser, Marcello Cinque, and Saurabh Bagchi.
\newblock {Learning from The Ones That Got Away: Detecting New Forms of
  Phishing Attacks}.
\newblock {\em IEEE Transactions on Dependable and Secure Computing}, 15(6),
  2018.

\bibitem{Hall2009}
Mark Hall, Eibe Frank, Geoffrey Holmes, Bernhard Pfahringer, Peter Reutemann,
  and Ian~H. Witten.
\newblock The weka data mining software: An update.
\newblock {\em SIGKDD Explor. Newsl.}, 11(1):10–18, November 2009.

\bibitem{hand2001simple}
David~J Hand and Robert~J Till.
\newblock A simple generalisation of the area under the roc curve for multiple
  class classification problems.
\newblock {\em Machine learning}, 45(2):171--186, 2001.

\bibitem{he_automl_2021}
Xin He, Kaiyong Zhao, and Xiaowen Chu.
\newblock {AutoML: {A} Survey of The State-of-the-art}.
\newblock {\em Knowledge-Based Systems}, 212:106622, January 2021.

\bibitem{ho2002complexity}
Tin~Kam Ho and Mitra Basu.
\newblock {Complexity Measures of Supervised Classification Problems}.
\newblock {\em IEEE Transactions on Pattern Analysis and Machine Intelligence},
  24(3):289--300, 2002.

\bibitem{ho2006measures}
Tin~Kam Ho, Mitra Basu, and Martin Hiu~Chung Law.
\newblock {Measures of Geometrical Complexity in Classification Problems}.
\newblock In {\em Data Complexity in Pattern Recognition}, pages 1--23.
  Springer, 2006.

\bibitem{Hong2012}
Jason Hong.
\newblock The state of phishing attacks.
\newblock {\em Commun. ACM}, 55(1):74--81, January 2012.

\bibitem{irsoy2017unsupervised}
Ozan Irsoy and Ethem Alpayd{\i}n.
\newblock Unsupervised feature extraction with autoencoder trees.
\newblock {\em Neurocomputing}, 258:63--73, 2017.

\bibitem{karras2019style}
Tero Karras, Samuli Laine, and Timo Aila.
\newblock A style-based generator architecture for generative adversarial
  networks.
\newblock In {\em Proceedings of the IEEE/CVF Conference on Computer Vision and
  Pattern Recognition}, pages 4401--4410, 2019.

\bibitem{komer2014hyperopt}
Brent Komer, James Bergstra, and Chris Eliasmith.
\newblock {Hyperopt-sklearn: Automatic Hyperparameter Configuration for
  Scikit-Learn}.
\newblock In {\em {Proc. SciPy}}, 2014.

\bibitem{kotthoff_auto-weka_2017}
Lars Kotthoff, Chris Thornton, Holger~H. Hoos, Frank Hutter, and Kevin
  Leyton-Brown.
\newblock Auto-{WEKA} 2.0: {Automatic} model selection and hyperparameter
  optimization in {WEKA}.
\newblock {\em Journal of Machine Learning Research}, 18(25):1--5, 2017.

\bibitem{krishnan2017boostclean}
Sanjay Krishnan, Michael~J Franklin, Ken Goldberg, and Eugene Wu.
\newblock Boostclean: Automated error detection and repair for machine
  learning.
\newblock {\em arXiv preprint arXiv:1711.01299}, 2017.

\bibitem{krishnan2019alphaclean}
Sanjay Krishnan and Eugene Wu.
\newblock Alphaclean: Automatic generation of data cleaning pipelines.
\newblock {\em arXiv preprint arXiv:1904.11827}, 2019.

\bibitem{lecun1998gradient}
Yann LeCun, L{\'e}on Bottou, Yoshua Bengio, and Patrick Haffner.
\newblock Gradient-based learning applied to document recognition.
\newblock {\em Proceedings of the IEEE}, 86(11):2278--2324, 1998.

\bibitem{H2OAutoML20}
Erin LeDell and Sebastien Poirier.
\newblock {H2O AutoML: Scalable Automatic Machine Learning}.
\newblock {\em 7th ICML Workshop on Automated Machine Learning (AutoML)}, July
  2020.

\bibitem{lin2019online}
Chen Lin, Minghao Guo, Chuming Li, Xin Yuan, Wei Wu, Junjie Yan, Dahua Lin, and
  Wanli Ouyang.
\newblock Online hyper-parameter learning for auto-augmentation strategy.
\newblock In {\em Proceedings of the IEEE/CVF International Conference on
  Computer Vision}, pages 6579--6588, 2019.

\bibitem{lingchen2020uniformaugment}
Tom~Ching LingChen, Ava Khonsari, Amirreza Lashkari, Mina~Rafi Nazari,
  Jaspreet~Singh Sambee, and Mario~A Nascimento.
\newblock Uniformaugment: A search-free probabilistic data augmentation
  approach.
\newblock {\em arXiv preprint arXiv:2003.14348}, 2020.

\bibitem{lorena2019complex}
Ana~C Lorena, Lu{\'\i}s~PF Garcia, Jens Lehmann, Marcilio~CP Souto, and Tin~Kam
  Ho.
\newblock {How Complex is Your Classification Problem? A Survey on Measuring
  Classification Complexity}.
\newblock {\em ACM Computing Surveys (CSUR)}, 52(5):1--34, 2019.

\bibitem{ma2009identifying}
Justin Ma, Lawrence~K Saul, Stefan Savage, and Geoffrey~M Voelker.
\newblock {Identifying Suspicious URLs: An Application of Large-Scale Online
  Learning}.
\newblock In {\em Proceedings of the 26th Annual International Conference on
  Machine Learning}, 2009.

\bibitem{mamun2016detecting}
Mohammad Saiful~Islam Mamun, Mohammad~Ahmad Rathore, Arash~Habibi Lashkari,
  Natalia Stakhanova, and Ali~A Ghorbani.
\newblock {Detecting Malicious URLs Using Lexical Analysis}.
\newblock In {\em International Conference on Network and System Security}.
  Springer, 2016.

\bibitem{marchal2014phishstorm}
Samuel Marchal, J{\'e}r{\^o}me Fran{\c{c}}ois, Radu State, and Thomas Engel.
\newblock {PhishStorm: Detecting Phishing with Streaming Analytics}.
\newblock {\em IEEE Transactions on Network and Service Management}, 11(4),
  2014.

\bibitem{meng2017relational}
Qinxue Meng, Daniel Catchpoole, David Skillicom, and Paul~J Kennedy.
\newblock Relational autoencoder for feature extraction.
\newblock In {\em 2017 International Joint Conference on Neural Networks
  (IJCNN)}, pages 364--371. IEEE, 2017.

\bibitem{mikolajczyk2018data}
Agnieszka Miko{\l}ajczyk and Micha{\l} Grochowski.
\newblock Data augmentation for improving deep learning in image classification
  problem.
\newblock In {\em 2018 international interdisciplinary PhD workshop (IIPhDW)},
  pages 117--122. IEEE, 2018.

\bibitem{mikolajczyk2019style}
Agnieszka Miko{\l}ajczyk and Micha{\l} Grochowski.
\newblock Style transfer-based image synthesis as an efficient regularization
  technique in deep learning.
\newblock In {\em 2019 24th International Conference on Methods and Models in
  Automation and Robotics (MMAR)}, pages 42--47. IEEE, 2019.

\bibitem{mohammad2012assessment}
Rami~M Mohammad, Fadi Thabtah, and Lee McCluskey.
\newblock {An Assessment of Features Related to Phishing Websites Using an
  Automated Technique}.
\newblock In {\em 2012 International Conference for Internet Technology and
  Secured Transactions}. IEEE, 2012.

\bibitem{oh2018learning}
Tae-Hyun Oh, Ronnachai Jaroensri, Changil Kim, Mohamed Elgharib, Fr'edo Durand,
  William~T Freeman, and Wojciech Matusik.
\newblock Learning-based video motion magnification.
\newblock In {\em Proceedings of the European Conference on Computer Vision
  (ECCV)}, pages 633--648, 2018.

\bibitem{olson2016}
Randal~S. Olson, Nathan Bartley, Ryan~J. Urbanowicz, and Jason~H. Moore.
\newblock Evaluation of a tree-based pipeline optimization tool for automating
  data science.
\newblock In {\em Proceedings of the Genetic and Evolutionary Computation
  Conference 2016}, GECCO '16, page 485–492, New York, NY, USA, 2016.
  Association for Computing Machinery.

\bibitem{OlsonGECCO2016}
Randal~S. Olson, Nathan Bartley, Ryan~J. Urbanowicz, and Jason~H. Moore.
\newblock {Evaluation of a Tree-based Pipeline Optimization Tool for Automating
  Data Science}.
\newblock In {\em Proceedings of the Genetic and Evolutionary Computation
  Conference 2016}, GECCO '16. ACM, 2016.

\bibitem{dcol2010}
A~Orriols-Puig, Núria Macià, and Tin Ho.
\newblock {DCoL: Data Complexity Library in C++ (Documentation)}, 2010.

\bibitem{scikit-learn}
F.~Pedregosa, G.~Varoquaux, A.~Gramfort, V.~Michel, B.~Thirion, O.~Grisel,
  M.~Blondel, P.~Prettenhofer, R.~Weiss, V.~Dubourg, J.~Vanderplas, A.~Passos,
  D.~Cournapeau, M.~Brucher, M.~Perrot, and E.~Duchesnay.
\newblock {Scikit-learn: Machine Learning in {P}ython}.
\newblock {\em Journal of Machine Learning Research}, 12:2825--2830, 2011.

\bibitem{ramanathan2012phishgillnet}
Venkatesh Ramanathan and Harry Wechsler.
\newblock {PhishGILLNET—Phishing Detection Methodology Using Probabilistic
  Latent Semantic Analysis, AdaBoost, and Co-training}.
\newblock {\em EURASIP Journal on Information Security}, 2012(1), 2012.

\bibitem{ruiz2018learning}
Nataniel Ruiz, Samuel Schulter, and Manmohan Chandraker.
\newblock Learning to simulate.
\newblock {\em arXiv preprint arXiv:1810.02513}, 2018.

\bibitem{sondhi2009feature}
Parikshit Sondhi.
\newblock Feature construction methods: a survey.
\newblock {\em sifaka. cs. uiuc. edu}, 69:70--71, 2009.

\bibitem{truong_towards_2019}
Anh Truong, Austin Walters, Jeremy Goodsitt, Keegan Hines, C.~Bayan Bruss, and
  Reza Farivar.
\newblock Towards {Automated} {Machine} {Learning}: {Evaluation} and
  {Comparison} of {AutoML} {Approaches} and {Tools}.
\newblock {\em 2019 IEEE 31st International Conference on Tools with Artificial
  Intelligence (ICTAI)}, pages 1471--1479, November 2019.
\newblock arXiv: 1908.05557.

\bibitem{vafaie1998evolutionary}
Haleh Vafaie and Kenneth De~Jong.
\newblock Evolutionary feature space transformation.
\newblock In {\em Feature Extraction, Construction and Selection}, pages
  307--323. Springer, 1998.

\bibitem{verma2013semantic}
Rakesh Verma and Nabil Hossain.
\newblock {Semantic Feature Selection for Text with Application to Phishing
  Email Detection}.
\newblock In {\em International Conference on Information Security and
  Cryptology}. Springer, 2013.

\bibitem{verma2019data}
Rakesh~M Verma, Victor Zeng, and Houtan Faridi.
\newblock {Data Quality for Security Challenges: Case Studies of Phishing,
  Malware and Intrusion Detection Datasets}.
\newblock In {\em Proceedings of the 2019 ACM SIGSAC Conference on Computer and
  Communications Security}, 2019.

\bibitem{waring_automated_2020}
J.~Waring, C.~Lindvall, and R.~Umeton.
\newblock {Automated Machine Learning: {Review} of the State-of-the-Art and
  Opportunities for Healthcare}.
\newblock {\em Artificial Intelligence in Medicine}, 104, 2020.

\bibitem{wong2016understanding}
Sebastien~C Wong, Adam Gatt, Victor Stamatescu, and Mark~D McDonnell.
\newblock Understanding data augmentation for classification: when to warp?
\newblock In {\em 2016 international conference on digital image computing:
  techniques and applications (DICTA)}, pages 1--6. IEEE, 2016.

\bibitem{zheng1998comparison}
Zijian Zheng.
\newblock A comparison of constructing different types of new feature for
  decision tree learning.
\newblock In {\em Feature Extraction, Construction and Selection}, pages
  239--255. Springer, 1998.

\end{thebibliography}

\clearpage
\appendix

\section{Model Performances}
\label{app:model_perform}
In this section, we provide more details regarding the AutoML and traditional ML model performances, as shown in Table~\ref{tab:automl_performance} and Table~\ref{tab:non_automl_performance}.

% Please add the following required packages to your document preamble:
% \usepackage{multirow}
% \usepackage{graphicx}
\begin{table}[H]
\centering
\begin{adjustbox}{width=0.625\textwidth,center}
%\resizebox{0.942\columnwidth}{!}{%
\begin{tabular}{llccc}
\hline
\textbf{Task} & \textbf{Framework} & \textbf{Accuracy} & \textbf{AUC Score} & \textbf{Duration (sec)} \\ \hline
\multirow{6}{*}{\textbf{eml\_1a}} & AutoGluon        & \textbf{0.866} & 0.865          & 3492.86  \\
                                  & auto-sklearn     & 0.865          & 0.869          & 3598.4   \\
                                  & GAMA             & 0.749          & 0.865          & 3239.22  \\
                                  & H2OAutoML        & 0.865          & \textbf{0.871} & 2442.72  \\
                                  & hyperopt-sklearn & 0.865          & 0.73           & 381.92   \\
                                  & TPOT             & 0.86           & 0.834          & 3315.22  \\ \hline
\multirow{6}{*}{\textbf{eml\_1b}} & AutoGluon        & \textbf{0.908} & \textbf{0.913} & 3525.68  \\
                                  & auto-sklearn     & 0.895          & \textbf{0.913} & 3601.26  \\
                                  & GAMA             & \textbf{0.908} & 0.908          & 3239.44  \\
                                  & H2OAutoML        & 0.899          & 0.906          & 2602.88  \\
                                  & hyperopt-sklearn & 0.887          & 0.821          & 429.83   \\
                                  & TPOT             & 0.904          & 0.908          & 3613.41  \\ \hline
\multirow{6}{*}{\textbf{eml\_1c}} & AutoGluon        & \textbf{0.965} & 0.967          & 3275.3   \\
                                  & auto-sklearn     & 0.933          & 0.964          & 3601.11  \\
                                  & GAMA             & 0.946          & 0.967          & 3241.95  \\
                                  & H2OAutoML        & 0.942          & \textbf{0.975} & 3289.61  \\
                                  & hyperopt-sklearn & 0.942          & 0.891          & 343.08   \\
                                  & TPOT             & 0.938          & 0.965          & 3047.91  \\ \hline
\multirow{6}{*}{\textbf{eml\_1d}} & AutoGluon        & \textbf{0.923} & \textbf{0.952} & 3527.28  \\
                                  & auto-sklearn     & 0.916          & 0.948          & 3598.93  \\
                                  & GAMA             & 0.92           & 0.947          & 3239.68  \\
                                  & H2OAutoML        & 0.899          & 0.946          & 2912.53  \\
                                  & hyperopt-sklearn & 0.897          & 0.871          & 31.57    \\
                                  & TPOT             & 0.915          & 0.947          & 3619.78  \\ \hline
\multirow{6}{*}{\textbf{eml\_1e}} & AutoGluon        & 0.896          & \textbf{0.901} & 3491.24  \\
                                  & auto-sklearn     & 0.897          & 0.876          & 3599.78  \\
                                  & GAMA             & 0.894          & 0.873          & 3240.91  \\
                                  & H2OAutoML        & \textbf{0.901} & 0.886          & 2621.21  \\
                                  & hyperopt-sklearn & 0.872          & 0.802          & 413.54   \\
                                  & TPOT             & 0.86           & 0.87           & 3418.89  \\ \hline
\multirow{6}{*}{\textbf{eml\_2}}  & AutoGluon        & 0.971          & 0.984          & 3461.46  \\
                                  & auto-sklearn     & 0.97           & \textbf{0.985} & 3600.33  \\
                                  & GAMA             & 0.64           & 0.984          & 3239.53  \\
                                  & H2OAutoML        & \textbf{0.972} & 0.984          & 2483.41  \\
                                  & hyperopt-sklearn & \textbf{0.972} & 0.967          & 879.84   \\
                                  & TPOT             & 0.824          & 0.984          & 3610.52  \\ \hline
\multirow{6}{*}{\textbf{url\_1}}  & AutoGluon        & nan            & nan            & 12113.62 \\
                                  & auto-sklearn     & 0.969          & \textbf{0.993} & 3596.35  \\
                                  & GAMA             & 0.97           & \textbf{0.993} & 3305.78  \\
                                  & H2OAutoML        & 0.964          & \textbf{0.993} & 3399.22  \\
                                  & hyperopt-sklearn & 0.931          & 0.916          & 3660.97  \\
                                  & TPOT             & \textbf{0.972} & 0.985          & 5274.93  \\ \hline
\multirow{6}{*}{\textbf{url\_2}}  & AutoGluon        & \textbf{0.961} & \textbf{0.994} & 2912.35  \\
                                  & auto-sklearn     & 0.952          & 0.991          & 3602.72  \\
                                  & GAMA             & 0.952          & 0.991          & 3257.05  \\
                                  & H2OAutoML        & 0.953          & 0.992          & 3281.59  \\
                                  & hyperopt-sklearn & 0.946          & 0.946          & 1812.67  \\
                                  & TPOT             & 0.952          & 0.991          & 3764.12  \\ \hline
\multirow{6}{*}{\textbf{url\_3}}  & AutoGluon        & \textbf{0.961} & \textbf{0.99}  & 3178.55  \\
                                  & auto-sklearn     & 0.952          & 0.986          & 3600.84  \\
                                  & GAMA             & 0.952          & 0.986          & 3272.36  \\
                                  & H2OAutoML        & 0.953          & 0.988          & 3375.62  \\
                                  & hyperopt-sklearn & 0.949          & 0.974          & 1094.88  \\
                                  & TPOT             & 0.952          & 0.986          & 3767.93  \\ \hline
\multirow{6}{*}{\textbf{web\_1}}  & AutoGluon        & 0.942          & 0.987          & 3438.56  \\
                                  & auto-sklearn     & 0.941          & 0.988          & 3599.88  \\
                                  & GAMA             & 0.945          & 0.989          & 3240.61  \\
                                  & H2OAutoML        & \textbf{0.947} & \textbf{0.99}  & 3199.18  \\
                                  & hyperopt-sklearn & 0.937          & 0.935          & 417.4    \\
                                  & TPOT             & 0.939          & 0.89           & 3766.47  \\ \bottomrule 
\end{tabular}%
%}
\end{adjustbox}
\caption{AutoML-based Model Performance}
\label{tab:automl_performance}
\end{table}

% Please add the following required packages to your document preamble:
% \usepackage{multirow}
% \usepackage{graphicx}
\begin{table}[H]
\centering
\begin{adjustbox}{width=0.625\textwidth,center}
%\resizebox{\columnwidth}{!}{%
\begin{tabular}{llccc}
\hline
\multicolumn{1}{c}{\textbf{Task}} & \multicolumn{1}{l}{\textbf{Framework}} & \multicolumn{1}{c}{\textbf{Accuracy}} & \textbf{AUC Score} & \textbf{Duration (sec)} \\ \hline
\multirow{7}{*}{\textbf{eml\_1a}} & Logistic Regression   & \textbf{0.749} & 0.473          & 1.641    \\
                                  & SVM                   & \textbf{0.749} & 0.473          & 2.214    \\
                                  & KNN                   & \textbf{0.749} & 0.399          & 2.179    \\
                                  & Decision tree         & \textbf{0.749} & 0.72           & 1.499    \\
                                  & Random forest         & \textbf{0.749} & 0.72           & 7.764    \\
                                  & Multilayer perceptron & \textbf{0.749} & 0.473          & 18.032   \\
                                  & Gaussian NB           & \textbf{0.749} & \textbf{0.736} & 0.001    \\ \hline
\multirow{7}{*}{\textbf{eml\_1b}} & Logistic Regression   & 0.883          & 0.862          & 2.709    \\
                                  & SVM                   & 0.834          & 0.828          & 3.608    \\
                                  & KNN                   & 0.752          & 0.642          & 2.195    \\
                                  & Decision tree         & 0.896          & 0.86           & 1.777    \\
                                  & Random forest         & \textbf{0.906} & \textbf{0.908} & 13.047   \\
                                  & Multilayer perceptron & 0.749          & 0.304          & 23.342   \\
                                  & Gaussian NB           & 0.863          & 0.849          & 0.002    \\ \hline
\multirow{7}{*}{\textbf{eml\_1c}} & Logistic Regression   & 0.945          & 0.973          & 10.8     \\
                                  & SVM                   & 0.943          & \textbf{0.983} & 28.961   \\
                                  & KNN                   & 0.775          & 0.765          & 29.746   \\
                                  & Decision tree         & 0.933          & 0.888          & 10.87    \\
                                  & Random forest         & 0.948          & 0.96           & 191.959  \\
                                  & Multilayer perceptron & \textbf{0.951} & 0.976          & 44.91    \\
                                  & Gaussian NB           & 0.702          & 0.777          & 0.045    \\ \hline
\multirow{7}{*}{\textbf{eml\_1d}} & Logistic Regression   & 0.909          & 0.934          & 1.912    \\
                                  & SVM                   & \textbf{0.917} & 0.937          & 2.586    \\
                                  & KNN                   & 0.903          & 0.919          & 2.963    \\
                                  & Decision tree         & 0.897          & 0.894          & 1.801    \\
                                  & Random forest         & 0.916          & \textbf{0.95}  & 19.209   \\
                                  & Multilayer perceptron & 0.749          & 0.934          & 19.411   \\
                                  & Gaussian NB           & 0.839          & 0.915          & 0.001    \\ \hline
\multirow{7}{*}{\textbf{eml\_1e}} & Logistic Regression   & 0.892          & 0.856          & 7.154    \\
                                  & SVM                   & 0.879          & 0.86           & 39.138   \\
                                  & KNN                   & 0.868          & 0.83           & 91.487   \\
                                  & Decision tree         & 0.845          & 0.766          & 3.076    \\
                                  & Random forest         & \textbf{0.901} & 0.847          & 24.441   \\
                                  & Multilayer perceptron & 0.893          & \textbf{0.874} & 36.772   \\
                                  & Gaussian NB           & 0.829          & 0.841          & 0.044    \\ \hline
\multirow{7}{*}{\textbf{eml\_2}}  & Logistic Regression   & 0.603          & 0.598          & 1.711    \\
                                  & SVM                   & 0.397          & 0.402          & 14.276   \\
                                  & KNN                   & 0.707          & 0.694          & 3.346    \\
                                  & Decision tree         & \textbf{0.741} & \textbf{0.752} & 1.647    \\
                                  & Random forest         & \textbf{0.741} & \textbf{0.752} & 11.034   \\
                                  & Multilayer perceptron & 0.603          & 0.5            & 36.178   \\
                                  & Gaussian NB           & 0.383          & 0.466          & 0.003    \\ \hline
\multirow{7}{*}{\textbf{url\_1}}  & Logistic Regression   & 0.975          & 0.995          & 473.486  \\
                                  & SVM                   & 0.788          & 0.88           & 5031.541 \\
                                  & KNN                   & 0.934          & 0.975          & 18751.06 \\
                                  & Decision tree         & 0.943          & 0.962          & 3847.875 \\
                                  & Random forest         & 0.971          & 0.99           & 120065.8 \\
                                  & Multilayer perceptron & \textbf{0.977} & \textbf{0.997} & 942.103  \\
                                  & Gaussian NB           & 0.731          & 0.769          & 0.773    \\ \hline
\multirow{7}{*}{\textbf{url\_2}}  & Logistic Regression   & 0.815          & 0.903          & 25.737   \\
                                  & SVM                   & 0.733          & 0.214          & 298.13   \\
                                  & KNN                   & 0.911          & 0.968          & 79.309   \\
                                  & Decision tree         & 0.9            & 0.959          & 11.277   \\
                                  & Random forest         & \textbf{0.955} & \textbf{0.992} & 279.35   \\
                                  & Multilayer perceptron & 0.511          & 0.365          & 282.544  \\
                                  & Gaussian NB           & 0.747          & 0.769          & 0.017    \\ \hline
\multirow{7}{*}{\textbf{url\_3}}  & Logistic Regression   & \textbf{0.94}  & 0.921          & 134.057  \\
                                  & SVM                   & \textbf{0.94}  & 0.89           & 184.147  \\
                                  & KNN                   & \textbf{0.94}  & 0.938          & 40.508   \\
                                  & Decision tree         & \textbf{0.94}  & 0.917          & 10.184   \\
                                  & Random forest         & \textbf{0.94}  & \textbf{0.981} & 233.451  \\
                                  & Multilayer perceptron & \textbf{0.94}  & 0.742          & 81.508   \\
                                  & Gaussian NB           & \textbf{0.94}  & 0.778          & 0.028    \\ \hline
\multirow{7}{*}{\textbf{web\_1}}  & Logistic Regression   & 0.923          & 0.974          & 3.601    \\
                                  & SVM                   & 0.659          & 0.756          & 15.978   \\
                                  & KNN                   & 0.913          & 0.966          & 51.269   \\
                                  & Decision tree         & 0.899          & 0.947          & 2.183    \\
                                  & Random forest         & \textbf{0.934} & \textbf{0.98}  & 20.192   \\
                                  & Multilayer perceptron & 0.919          & 0.973          & 47.325   \\
                                  & Gaussian NB           & 0.588          & 0.965          & 0.005    \\ \bottomrule
\end{tabular}%
%}
\end{adjustbox}
\caption{Manually Developed (non-AutoML) Model Performance}
\label{tab:non_automl_performance}
\end{table}

\section{Correlation Analysis}
\label{app:correlation_analysis}

In this section, we provide results on the analysis of correlation between the complexity measures and the AutoML performances (Table~\ref{tab:correlation_automl}), and correlation between the complexity measures and the classification performance gain when using AutoML frameworks (Table~\ref{tab:correlation_gain}).

% Please add the following required packages to your document preamble:
% \usepackage{booktabs}
% \usepackage{graphicx}
\begin{table}[H]
\centering
\begin{adjustbox}{width=0.725\textwidth,center}
%\resizebox{\columnwidth}{!}{%
\begin{tabular}{@{}cccc@{}}
\toprule
\textbf{Complexity Measure} & \textbf{Performance Metric} & \textbf{Correlation} & \textbf{p-value} \\ \midrule
F1  & Accuracy  & 0.151515 & 0.676065          \\
F1  & F1 Score  & 0.187879 & 0.603218          \\
F1  & AUC Score & -0.21212 & 0.556306          \\
F1v & Accuracy  & -0.24848 & 0.488776          \\
F1v & F1 Score  & -0.32121 & 0.365468          \\
F1v & AUC Score & -0.55152 & 0.098401          \\
F2  & Accuracy  & 0.006465 & 0.985858          \\
F2  & F1 Score  & -0.12284 & 0.735313          \\
F2  & AUC Score & 0.071116 & 0.845218          \\
F3  & Accuracy  & 0.515152 & 0.127553          \\
F3  & F1 Score  & 0.563636 & 0.089724          \\
F3  & AUC Score & 0.260606 & 0.467089          \\
F4  & Accuracy  & 0.648485 & \textbf{0.04254}  \\
F4  & F1 Score  & 0.587879 & 0.073878          \\
F4  & AUC Score & 0.284848 & 0.425038          \\
L1  & Accuracy  & 0.418182 & 0.229113          \\
L1  & F1 Score  & 0.393939 & 0.259998          \\
L1  & AUC Score & 0.139394 & 0.700932          \\
L2  & Accuracy  & -0.32121 & 0.365468          \\
L2  & F1 Score  & -0.17576 & 0.627188          \\
L2  & AUC Score & -0.12727 & 0.726057          \\
L3  & Accuracy  & -0.69697 & \textbf{0.025097} \\
L3  & F1 Score  & -0.68485 & \textbf{0.028883} \\
L3  & AUC Score & -0.7697  & \textbf{0.009222} \\
N1  & Accuracy  & -0.11515 & 0.75142           \\
N1  & F1 Score  & -0.04242 & 0.907364          \\
N1  & AUC Score & 0.29697  & 0.404702          \\
N2  & Accuracy  & 0.090909 & 0.802772          \\
N2  & F1 Score  & -0.15152 & 0.676065          \\
N2  & AUC Score & -0.29697 & 0.404702          \\
N3  & Accuracy  & -0.10303 & 0.776998          \\
N3  & F1 Score  & -0.06667 & 0.854813          \\
N3  & AUC Score & 0.260606 & 0.467089          \\
N4  & Accuracy  & -0.28485 & 0.425038          \\
N4  & F1 Score  & -0.13939 & 0.700932          \\
N4  & AUC Score & 0.10303  & 0.776998          \\
T1  & Accuracy  & 0.830303 & \textbf{0.00294}  \\
T1  & F1 Score  & 0.781818 & \textbf{0.007547} \\
T1  & AUC Score & 0.587879 & 0.073878          \\
T2  & Accuracy  & 0.127273 & 0.726057          \\
T2  & F1 Score  & 0.151515 & 0.676065          \\
T2  & AUC Score & 0.2      & 0.579584          \\
C1  & Accuracy  & -0.5366  & 0.109784          \\
C1  & F1 Score  & -0.69176 & \textbf{0.026678} \\
C1  & AUC Score & -0.834   & \textbf{0.002705} \\
C2  & Accuracy  & -0.57539 & 0.081792          \\
C2  & F1 Score  & -0.73055 & \textbf{0.016409} \\
C2  & AUC Score & -0.88572 & \textbf{0.000649} \\ \bottomrule
\end{tabular}%
%}
\end{adjustbox}
\caption{Correlation between Complexity Measure and AutoML Performance}
\label{tab:correlation_automl}
\end{table}

\newpage

% Please add the following required packages to your document preamble:
% \usepackage{booktabs}
% \usepackage{graphicx}
\begin{table}[H]
\centering
%\resizebox{\columnwidth}{!}{%
\begin{adjustbox}{width=0.725\textwidth,center}
\begin{tabular}{@{}cccc@{}}
\toprule
\textbf{Complexity Measure} & \textbf{Performance Gain} & \textbf{Correlation} & \textbf{p-value} \\ \midrule
F1  & Accuracy  & -0.21212 & 0.556306          \\
F1  & F1 Score  & -0.38182 & 0.276255          \\
F1  & AUC Score & -0.70909 & \textbf{0.021666} \\
F1v & Accuracy  & 0.272727 & 0.445838          \\
F1v & F1 Score  & 0.333333 & 0.346594          \\
F1v & AUC Score & 0.006061 & 0.986743          \\
F2  & Accuracy  & 0.536602 & 0.109784          \\
F2  & F1 Score  & 0.316789 & 0.37248           \\
F2  & AUC Score & 0.38144  & 0.276771          \\
F3  & Accuracy  & -0.62424 & 0.053718          \\
F3  & F1 Score  & -0.41818 & 0.229113          \\
F3  & AUC Score & -0.67273 & \textbf{0.033041} \\
F4  & Accuracy  & -0.70909 & \textbf{0.021666} \\
F4  & F1 Score  & -0.44242 & 0.200423          \\
F4  & AUC Score & -0.53939 & 0.107593          \\
L1  & Accuracy  & 0.357576 & 0.310376          \\
L1  & F1 Score  & -0.45455 & 0.186905          \\
L1  & AUC Score & -0.49091 & 0.149656          \\
L2  & Accuracy  & 0.575758 & 0.081553          \\
L2  & F1 Score  & 0.115152 & 0.75142           \\
L2  & AUC Score & 0.115152 & 0.75142           \\
L3  & Accuracy  & -0.16364 & 0.651477          \\
L3  & F1 Score  & 0.078788 & 0.828717          \\
L3  & AUC Score & -0.06667 & 0.854813          \\
N1  & Accuracy  & 0.563636 & 0.089724          \\
N1  & F1 Score  & 0.539394 & 0.107593          \\
N1  & AUC Score & 0.745455 & \textbf{0.01333}  \\
N2  & Accuracy  & -0.15152 & 0.676065          \\
N2  & F1 Score  & -0.47879 & 0.161523          \\
N2  & AUC Score & -0.33333 & 0.346594          \\
N3  & Accuracy  & 0.624242 & 0.053718          \\
N3  & F1 Score  & 0.490909 & 0.149656          \\
N3  & AUC Score & 0.721212 & \textbf{0.018573} \\
N4  & Accuracy  & 0.127273 & 0.726057          \\
N4  & F1 Score  & 0.612121 & 0.059972          \\
N4  & AUC Score & 0.769697 & \textbf{0.009222} \\
T1  & Accuracy  & 0.10303  & 0.776998          \\
T1  & F1 Score  & -0.57576 & 0.081553          \\
T1  & AUC Score & -0.38182 & 0.276255          \\
T2  & Accuracy  & 0.030303 & 0.933773          \\
T2  & F1 Score  & -0.4303  & 0.214492          \\
T2  & AUC Score & -0.26061 & 0.467089          \\
C1  & Accuracy  & -0.47195 & 0.168458          \\
C1  & F1 Score  & -0.0194  & 0.957589          \\
C1  & AUC Score & -0.16163 & 0.655533          \\
C2  & Accuracy  & -0.38144 & 0.276771          \\
C2  & F1 Score  & -0.05819 & 0.873149          \\
C2  & AUC Score & -0.21335 & 0.553971          \\ \bottomrule
\end{tabular}%
\end{adjustbox}
%}
\caption{Correlation between Complexity Measure and Performance Gain with AutoML}
\label{tab:correlation_gain}
\end{table}

\newpage

\end{document}